\documentclass[screen]{jair}

\usepackage{graphicx}
\usepackage{multirow}
\usepackage{multicol}
\usepackage{url}
\usepackage{caption}
\usepackage{subcaption}
\usepackage[table]{xcolor}

\setcopyright{cc}
\copyrightyear{2025}
\acmYear{2025}
\acmDOI{10.1613/jair.1.xxxxx}

\begin{document}

\title{A Comparative Study of Controlled Text Generation Systems Using Level-Playing-Field Evaluation Principles}

\author{Michela Lorandi}
\authornote{Corresponding Author.}
\email{michela.lorandi@mail.dcu.ie}
\orcid{0000-0002-6131-8763}
\affiliation{%
  \institution{Dublin City University}
  \city{Dublin}
  \country{Ireland}
}

\author{Anya Belz}
\orcid{0000-0002-0552-8096}
\email{anya.belz@dcu.ie}
\affiliation{%
  \institution{Dublin City University}
  \city{Dublin}
  \country{Ireland}}


\begin{abstract}
{\bf Background:} 
    Many different approaches to controlled text generation (CTG) have been proposed over recent years, but it is difficult to get a clear picture of which approach performs best, because different datasets and evaluation methods are used in each case to assess the control achieved.  
    
    {\bf Objectives:}
    Our aim in the work reported in this paper is to develop an approach to evaluation that enables us to comparatively evaluate different CTG systems in a manner that is both informative and fair to the individual systems.
    
    {\bf Methods:}
    We use a level-playing-field (LPF) approach to comparative evaluation where we (i) generate and process all system outputs in a standardised way, and (ii) apply a shared set of evaluation methods and datasets, selected based on those currently in use, in order to ensure fair evaluation.  
    
    {\bf Results:}
    When re-evaluated in this way, performance results for a representative set of current CTG systems differ substantially from originally reported results, in most cases for the worse. This highlights the importance of a shared standardised way of assessing controlled generation.
    
    {\bf Conclusions:}
    The discrepancies revealed by LPF evaluation demonstrate the urgent need for standardised, reproducible evaluation practices in CTG. Our results suggest that without such practices, published performance claims may substantially misrepresent true system capabilities.    

\end{abstract}



\received{20 February 2007}
\received[revised]{12 March 2009}
\received[accepted]{5 June 2009}

\maketitle

\section{Introduction}
\label{Introduction}
Over the past few years, a slew of controlled text generation techniques have been proposed, controlling properties of the generated text such as its sentiment \cite{kim-etal-2023-critic}, its topic(s) \cite{gu-etal-2022-improving}, what specific words appear in it \cite{carlsson-etal-2022-fine}, or its level of toxicity \cite{gu-etal-2022-improving}, among many other aspects. Yet no clear picture emerges as to which techniques work better overall, or even for specific tasks, mainly because papers use a variety of different datasets and evaluation methods for assessing the degree to which a method achieves control over the properties it is intended to control. 

For example, Krause et al. \cite{krause-etal-2021-gedi-generative}, Yue et al. \cite{yu-etal-2021-attribute-alignment} and Zhang and Song \cite{zhang-song-2022-discup} all report methods for controlling the (binary) sentiment of the generated text and use a sentiment classifier to assess the degree to which their systems achieve sentiment control. However, Krause et al. \cite{krause-etal-2021-gedi-generative} evaluate on the Bookcorpus test set, using RoBERTa fine-tuned on Stanford Sentiment Treebank 2 (SST-2)~\cite{socher-etal-2013-recursive} as the classifier, whereas Yu et al. \cite{yu-etal-2021-attribute-alignment} evaluate on PPLM Prompts, using BERT trained on IMDb Movie as the classifier, and Zhang and Song \cite{zhang-song-2022-discup} evaluate on OpenWebText, with DistilBERT fine-tuned on SST-2 as the classifier. In this situation, clearly it is not possible to determine which of these approaches works better for sentiment control. Moreover, if, as is often the case, a single classifier is used as the evaluation metric, then evaluation results obtained with it may be misleading, because one classifier can give very different results from another classifier on a given test set.

In this paper, we compare different Controlled Text Generation (CTG) systems adopting a {level-playing-field (LPF) approach} for fair comparative evaluation. This approach addresses how to select datasets, evaluation metrics and systems to compare, and how to obtain outputs from them for the selected datasets, in a way that is fair to the different systems, and yields a truer assessment of the performance of different CTG techniques.

We start in Section~\ref{sec:rel-res} below with a review of related work.
We then present the LPF approach (Section~\ref{sec:lpfe-design}), and apply it to systematically selected sets of systems that implement control over (i) binary sentiment, (ii) topic, and (iii) included keywords in the generated text (Section~\ref{sec:models}). We describe our experimental set up (Section~\ref{sec:exp-setup}), provide analysis of the evaluation metrics used (Section~\ref{sec:eval_meth_analysis}) and report results for comparative evaluation of the selected systems with LPF (Section~\ref{sec:results}). We further compare the evaluation results we obtained to those reported in the original papers (Section~\ref{sec:comp_ref_papers}), and assess systems in terms of efficiency (Section~\ref{sec:eff_res}).  Finally, we conclude with some discussion and limitations (Sections~\ref{sec:disc} and~\ref{sec:lim}) and final remarks (Section~\ref{sec:conclusion}).

All our code, generated texts and results can be found on GitHub.\footnote{\url{https://github.com/michelalorandi/comparative_study_of_ctg}}

\section{Related Research}
\label{sec:rel-res}

In this section, we focus on two types of previous research related to the work presented in this paper: (i) research that explicitly addresses issues of bias and comparability in evaluation methods in Controlled Text Generation (CTG) and beyond; (ii) comparative studies of the performance of different CTG systems more generally; and (iii) recent work on developing CTG systems. In the latter case, we focus on different types of approaches taken, including type of control and how implemented, with the aim of underpinning the systematic selection of papers for our comparative study which form a representative set of state-of-the-art CTG techniques and are described in some detail in Section~\ref{sec:models}.

Third-party comparative studies like the one we are reporting in this paper are not uncommon in NLP, because of the diversity of techniques proposed in different NLP subfields, and the lack of standard evaluation methods in many. \citet{he-etal-2023-blind} show that widely used metrics like BERTScore and MAUVE are insensitive to certain errors, such as truncation and positional discrepancies, which can lead to unfair evaluations in summarization and other tasks.

Similarly, \citet{sun-etal-2022-bertscore} systematically explore social biases in pre-trained language model-based metrics, such as BERTScore, showing that these metrics favour models aligned with dominant social narratives. They propose debiasing adapters as a way to improve fairness while preserving evaluation performance, highlighting the risks of relying solely on unadjusted metrics for model evaluation.

Additionally, different work explores the limitations of the means-based evaluation metrics commonly used in NLP. \citet{gehrmann2023repairing} survey the major obstacles in evaluating generated text, identifying inconsistencies in metric use, a lack of correlation with human judgements, and a general absence of standardised evaluation protocols. \citet{peyrard-etal-2021-better} argue that aggregating scores can obscure instance-level variations, resulting in evaluations that fail to fully capture the strengths and weaknesses of different systems. \citet{colombo_what_2022} propose a novel approach to NLP benchmarking by introducing a method that aggregates system performances across multiple tasks. Their method aims to provide a more robust and fair comparison framework for evaluating NLP systems.

Finally, \citet{colombo-etal-2023-glass} analyse the limitations of automatic evaluation metrics in Natural Language Generation (NLG), revealing that these metrics often correlate more with each other than with human judgments. This suggests that current automatic metrics may not adequately correlate with human evaluation, underscoring the need for more reliable assessment tools.

In their different ways, the contributions above all aim to make evaluation fairer to the systems assessed, and emphasize the need for strong and unbiased evaluation methods for a truer comparison of NLP systems.

Some comparative studies have focused on Data-to-Text generation, such as \citet{castro-ferreira-etal-2019-neural} and \citet{leng-etal-2020-controllable} who systematically compare neural pipeline architectures, end-to-end approaches, and different control strategies in data-to-text generation. On a different task, Araújo et al. \cite{ARAUJO20201078} investigate the performance of machine translation systems for multilingual sentence-level sentiment analysis. \citet{cunha2021cost} compare text classification systems, proposing a systematic selection process of techniques and a fair evaluation process over multiple metrics and datasets. We apply a similar approach but focus on CTG techniques and additionally use the same set of datasets and metrics for all techniques for a fair assessment.

Closely related to our work, \citet{sun-etal-2023-evaluating} conduct an extensive evaluation of LLMs prompting on different CTG tasks. Their evaluation includes both raw and instruction-tuned LLMs, showing that LLMs struggle to satisfy fine-grained hard constraints. While we include LLM prompting in our comparative study, we systematically select a diverse range of CTG techniques, enabling a broader and more comprehensive comparison.

CTG has gained increasing attention over recent years, and several surveys have been conducted to provide overviews of this growing field, each focusing on different aspects of CTG work. For example, \citet{prabhumoye2020exploring} focus on components that control the generation process, \citet{lorandi-belz-2023-control} propose a typology based on how control is implemented in techniques, \citet{zhang2023survey} analyse CTG techniques in terms of how the control signal works in the pretrained language model (PLM), and \citet{wang2024recent} analyse CTG techniques from the perspective of causality theory instead of relying only on statistical associations. 

Building on these surveys, here we provide an overview of current state-of-the-art CTG techniques, categorising them based on their control implementation methods and the types of control they enable.

Many studies propose techniques that rely on fine-tuning pretrained models to achieve controlled text generation. For instance, some works adopt prefix tuning, where lightweight task-specific vectors are prepended to the model's input to guide the generation process \cite{ding-etal-2023-maclasa,yu-etal-2021-attribute-alignment,qian-etal-2022-controllable}. Other studies focus on full model fine-tuning, where the entire pretrained model is adapted to the specific task, allowing for greater flexibility in achieving control objectives \cite{sun-etal-2021-iga,carlsson-etal-2022-fine}. While highly effective, full fine-tuning is also very resource-intensive, motivating the search for alternative, more efficient strategies.

Other approaches train a model from scratch for the desired task, using a variety of architectures, such as Transformers \cite{casas-etal-2020-syntax}, Diffusion models \cite{tang-etal-2023-diffusion}, and Generative Adversarial Networks (GANs) \cite{betti-etal-2020-controlled}. While training from scratch allows for highly specialised models, it is computationally expensive and less practical in many real-world scenarios compared to fine-tuning or prompting.

Prompting, by contrast, avoids modifying model parameters altogether by steering the generation process via carefully crafted input prompts. Most studies employ manual prompt design, where task-specific prompts are explicitly written to elicit the desired outputs \cite{chen-etal-2023-controllable,pu-demberg-2023-chatgpt,liu-etal-2022-multi}.

Other techniques that avoid modifying model parameters focus on manipulating the token distribution during the generation process. In some methods, an external model modifies the token distribution to align it with the desired attribute, thereby guiding the text generation \cite{madotto-etal-2020-plug,pascual-etal-2021-plug-play}. Other work proposes novel sampling procedures, such as gradient-based sampling \cite{kumar-etal-2022-gradient}, beam search refinement \cite{landsman-etal-2022-beamr}, and neurologically inspired sampling techniques \cite{lu-etal-2022-neurologic}, which can be applied to any language model. These approaches offer flexibility and adaptability across diverse models and tasks.

Finally, hybrid techniques combine elements from multiple approaches, such as model fine-tuning and token distribution modification. For instance, some techniques fine-tune an external model, which is then applied to steer the generation process by adjusting the next-token distribution \cite{liu-etal-2022-multi-attribute,fan-etal-2023-nano}. Other approaches involve fine-tuning multiple versions of the same model and integrating their output distributions to determine the final token distribution, thereby balancing precision and diversity in generation \cite{ma-etal-2023-focused,kim-etal-2023-critic}. These hybrid methods exemplify how combining strengths from different paradigms can address the limitations of individual approaches, further enriching the landscape of CTG techniques.

Section~\ref{sec:models} provides a detailed description of the specific CTG techniques included in this comparative study.

\section{Comparative Study Design}\label{sec:lpfe-design}

This study uses level-playing-field (LPF) principles to ensure a fair and informative comparative evaluation of CTG systems. The principles address potential biases arising from system selection, evaluation metric selection, test data variability, and generation/post-processing discrepancies. More specifically, they relate to the following key aspects:

\begin{itemize}
    \item \textbf{Systematic selection of CTG methods (Section~\ref{sec:sel-methods}):} We adopt a systematic approach to identifying a representative and unbiased set of CTG systems for evaluation. This ensures coverage of diverse control techniques while adhering to objective inclusion criteria.
    \item \textbf{Evaluation methods (Section~\ref{sec:lev_play_method}):} The choice of evaluation metrics can influence a system's performance, potentially favouring one system over another. To mitigate this, we use a standardised and diverse set of metrics that evaluate multiple performance dimensions. 
    \item \textbf{Selection of datasets (Section~\ref{sec:data}):} When evaluations are based on a narrow range of datasets or samples, results may not generalise well to broader applications. Additionally, using datasets that align too closely with specific systems can unfairly influence the outcomes. Multiple test datasets are chosen to provide a balanced set of assessment tasks with the objective to minimize of mitigating the influence of dataset-specific biases. 
    \item \textbf{Output generation and post-processing (Section~\ref{sec:obtaining-outputs}):} Outputs from all systems are generated and processed in a standardised way to eliminate spurious differences arising from variations in generation or post-processing protocols. 
\end{itemize}

\noindent The above principles aim to ensure that systems are evaluated on an equal footing (a \textit{level playing-field}), thereby enabling a truer performance comparison. The following subsections describe how we implemented the principles in the present study.

\subsection{Systematic selection of CTG methods}\label{sec:sel-methods}

A critical issue in comparative evaluation is the introduction of bias during the selection of systems to be evaluated. Bias can arise when the choice of systems is influenced by subjective preferences, inconsistent criteria, or inadequate representation of the diversity of approaches. 

\begin{figure}[t!]
    \centering
    \includegraphics[width=1\textwidth]{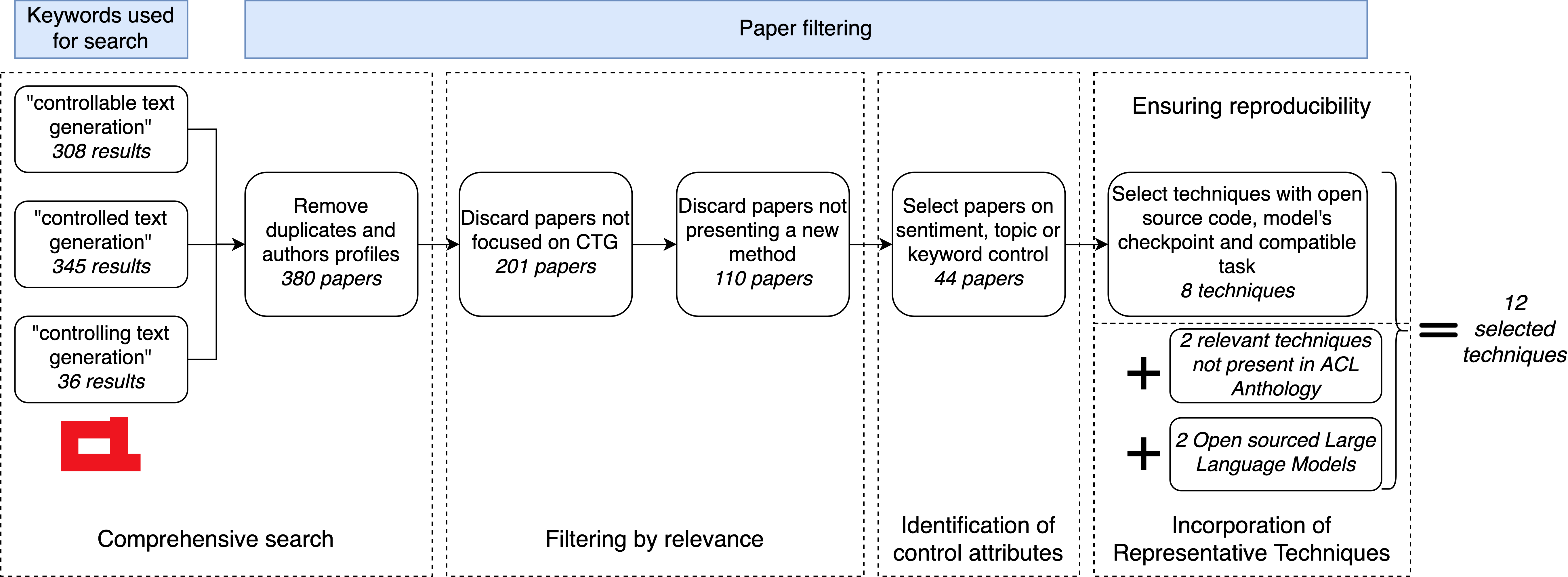}
    \caption{CTG techniques selection process from ACL Anthology to which we added 2 open-source LLMs and 2 relevant techniques not in the ACL Anthology.}
    \label{fig:sel_process}
\end{figure}

To address this, we adopt a systematic, multi-step process (Figure~\ref{fig:sel_process}) to ensure fairness and inclusivity in the selection of CTG systems. We apply a transparent and replicable process with objective criteria with the following main steps: comprehensive search, filtering by relevance, identification of control attributes, ensuring reproducibility, and incorporation of representative techniques.

\textbf{Comprehensive search.} We conducted a broad literature search of the ACL Anthology in early September 2023 using the keywords ``controllable text generation'', ``controlled text generation'' and ``controlling text generation'' to identify candidate CTG techniques.

\textbf{Filtering by relevance.} Papers not directly addressing CTG or lacking implementation details were excluded. After removing duplicates, authors' profiles, and non-paper resources, we were left with 380 papers. We then filtered out all papers not addressing CTG but contained references to CTG methods, resulting in 201 papers left. 
From these, we excluded papers that did not present a new model or control method (e.g.\ works that only presented new datasets), narrowing our selection to 110 papers.

\textbf{Identification of control attributes.} Next, we identified the most frequently used control attributes among the remaining papers, to enable us to select comparable systems for the LPF evaluations. We found that sentiment, topic, and keyword control were the most used attributes in the 110 papers, making them the most appropriate for our investigation. Specifically, we selected systems (44 papers) that implement techniques for control over (i) \textbf{binary sentiment}, (ii) \textbf{topic}, and (iii) \textbf{included keywords} in the generated output text (Section~\ref{sec:models}). 

\textbf{Filtering by code and model availability.} To ensure consistency and reproducibility, we further refined our selection to include only those systems for which open-source code and model checkpoints were available (10 systems). For topic control, we ensured a consistent set of topics by discarding 2 systems that did not have a direct mapping with the topics used in the other 8 papers. 

\textbf{Incorporation of representative techniques.} To the list of techniques to evaluate, we added two LLM-prompting techniques using open-source LLMs (Falcon 40B Instruct, and LLaMa2 70B Chat) to provide a point of comparison between general-purpose instruction-tuned LLMs and models specifically designed for CTG. Finally, we included two additional papers that met our selection criteria but were sourced from outside the ACL Anthology, as these systems were referenced and used as baselines in the majority of the selected papers.

This approach aims to ensure that the final set of selected systems represents a diverse and relevant sample of state-of-the-art CTG techniques while minimizing bias introduced during selection.

Table~\ref{tab:ctg_techniques} provides an overview of the final set of 12 selected CTG techniques, showing in each case the type of pretrained language model (PLM) incorporated, the task implemented, the types of control addressed, and datasets used for evaluation.  In some cases, no specific evaluation dataset is identified in the original papers. This occurs, for example, with general-purpose LLMs, which were not originally developed for CTG tasks and thus do not include CTG-specific evaluation datasets. In one case, although the system was evaluated, the paper does not disclose which dataset was used. For consistency in our study, all systems were evaluated on the same set of datasets, as described in Section~\ref{sec:data}.

\begin{table*}[ht!]
    \centering
    \small
    \setlength\tabcolsep{2pt} 
    \renewcommand{\arraystretch}{1.15}
    \begin{tabular}{|l|c|c|c|ccc|l|}
\hline
\multirow{2}{*}{\textbf{CTG Technique}} & \multirow{2}{*}{\textbf{Model}} & \multirow{2}{*}{\textbf{Task}} & \multirow{2}{*}{\textbf{C}} & \multicolumn{3}{c|}{\textbf{Attributes}} & \multirow{2}{*}{\textbf{Evaluation Datasets}} \\
 & & & & \textbf{S} & \textbf{T} & \textbf{K} & \\
\hline
\multicolumn{8}{|c|}{\textit{Complete Training}} \\
\hline
CTRL \cite{keskar2019ctrl} & Transf & FT &Si & & $\checkmark$ & & - \\
 \hline
\multicolumn{8}{|c|}{\textit{Model Fine-Tuning}} \\
\hline
C BART \cite{he-2021-parallel} & BART & FT &Si  & & & $\checkmark$ & One-Billion-Word dataset \cite{chelba2013one}, Yelp dataset \cite{Zhang2015CharacterlevelCN} \\
Multi CTG \cite{gu-etal-2022-distributional} & BERT & FT &M  & $\checkmark$ & $\checkmark$ & & PPLM Prompts \cite{dathathri2019plug} \\
Prior CTG \cite{gu-etal-2023-controllable} & GPT-2 M & FT &M  & $\checkmark$ & $\checkmark$ & & PPLM Prompts 
\\
\hline
\multicolumn{8}{|c|}{\textit{Modification of Token Distribution}} \\
\hline
CAT-PAW  \cite{gu-etal-2022-improving} & GPT-2 M & FT &Si  & $\checkmark$ & $\checkmark$ & & PPLM Prompts 
\\
PPLM \cite{dathathri2019plug} & GPT-2 M & FT &M  & $\checkmark$ & $\checkmark$ & & PPLM Prompts 
\\
\hline
\multicolumn{8}{|c|}{\textit{Prompting}} \\
\hline
Falcon 40B Instruct \cite{falcon40B} & Falcon & FT &M  & $\checkmark$ & $\checkmark$ & $\checkmark$  & - \\
LLaMa2 70B chat \cite{touvron2023llama} & LLaMa2 & FT &M  & $\checkmark$ & $\checkmark$ & $\checkmark$  & - \\
 \hline
\multicolumn{8}{|c|}{\textit{Hybrid}} \\
\hline
GeDi \cite{krause-etal-2021-gedi-generative} & GPT-2 XL & FT &Si & $\checkmark$ & $\checkmark$ & & Bookcorpus (S) \cite{zhu2015aligning}, AGnews (T) \cite{Zhang2015CharacterlevelCN} \\
DisCup \cite{zhang-song-2022-discup} & GPT-2 L & FT &Si  & $\checkmark$ & & & OpenWebText corpus 
\\
DExperts \cite{liu-etal-2021-dexperts} & GPT-2 L & FT &Si  & $\checkmark$ & & & OpenWebText corpus 
\\
BOLT \cite{liu-etal-2023-bolt} & GPT-2 L & FT &Si  & $\checkmark$ & & $\checkmark$  & PPLM Prompts (S, K)
, CommonGen (K) \cite{lin-etal-2020-commongen} \\
\hline
\end{tabular}
    \caption{Overview of the selected CTG techniques. FT=Free Text Generation, C=Control, Si=Single-attribute control, M=Multiple-attribute control, S=Sentiment, T=Topic, K=Keywords.} 
    \label{tab:ctg_techniques}
\end{table*}

\subsection{Evaluation methods}
\label{sec:lev_play_method}

Bias can arise from the choice of evaluation metrics and how these metrics are implemented. A different set of metrics will almost certainly produce differences in system rankings. For example, using different \textit{single} classifiers, e.g.\ for sentiment classification to test whether the generated text does have the targeted sentiment, can produce very different evaluation scores for a given sentiment-control system, and different rankings for a set of such systems. If a classifier is moreover trained on a single dataset, the bias is amplified. 

In our LPF approach, we combine standardisation of metric implementation, execution and application with diversity of selected metrics. The former is to ensure that arbitrary differences in what should be exactly the same metric do not affect results. The latter is to provide a balanced picture of quality where multiple quality criteria are applicable. While we standardise output preprocessing, evaluation metrics and datasets to ensure systems are evaluated under the same conditions, we also address potential biases introduced by how evaluation metrics are implemented or how models are trained. For example, instead of relying on a single classifier or model, we use multiple classifiers, models, or approaches to compute the metrics. This reduces the risk of biases from over-specialisation or training for a specific purpose.

To implement this approach, we propose a multi-faceted evaluation framework designed to provide a comprehensive assessment of system performance. The framework evaluates three key aspects of system outputs: \textit{Diversity}, \textit{Fluency}, and \textit{Control Effectiveness (CE)}.

\subsubsection*{General evaluation aspects} All sets of system outputs are evaluated with (i) Distinct-n \cite{li2015diversity} (n=1, 2, 3), which computes the proportion of unique n-grams, and (ii) Syntactic Log-Odds Ratio (SLOR)~\cite{kann-etal-2018-sentence}, which computes the log-probability of a sentence normalised by unigram log-probability and length (for more details see Section~\ref{sec:ppl_vs_slor}).
System-level Distinct-n is calculated as the average control-level Distinct-n, while system-level SLOR is calculated as the average SLOR using GPT-2 XL \cite{radford2019language} and BLOOM 1B7 \cite{scao2022bloom}.

\subsubsection*{Evaluation of Control Effectiveness} CE computes the percentage of texts identified as displaying the desired control attribute according to a given measure. To avoid biases from using just one classifier to assess CE, we use three different classifiers and compute their average as the final overall score. Section~\ref{sec:avg_vs_vote} provides a comprehensive analysis of average and majority vote versions of CE assessment. The classifiers are chosen based on two main criteria: first, the availability of classifiers used in the selected CTG techniques, and second, a search on the HuggingFace ecosystem using the ``Text Classification'' filter, sorting models by most downloads. To avoid bias arising from the usage of only one model architecture, we try to select classifiers based on different architectures where possible. We do not retrain or finetune our classifiers to prevent introducing new biases into CE evaluation.

\textbf{Sentiment control evaluation}. We apply three sentiment classifiers to system output texts and compute the proportion of times the result of classification matches the intended output sentiment for each classifier and then average those proportions to give an overall Sentiment Control Effectiveness score. The three classifiers we use for this computation are DistilBERT\footnote{\url{https://huggingface.co/distilbert/distilbert-base-uncased-finetuned-sst-2-english}} and T5 fine-tuned on SST-2,\footnote{\url{https://huggingface.co/michelecafagna26/t5-base-finetuned-sst2-sentiment}} and DeBERTa fine-tuned on Yelp \cite{gu-etal-2023-controllable}.

\textbf{Topic control evaluation}. Topic Control Effectiveness is computed similarly using three topic classifiers, namely DistilBERT,\footnote{\url{https://huggingface.co/textattack/distilbert-base-uncased-ag-news}} BERT,\footnote{\url{https://huggingface.co/fabriceyhc/bert-base-uncased-ag_news}} and DeBERTa \cite{gu-etal-2023-controllable}, all fine-tuned on AG-News. DeBERTa is fine-tuned as a set of binary classifiers, one per topic, and appears to have been used previously by Gu et al. \cite{gu-etal-2023-controllable} by first picking the topic and then deciding class membership by threshold based only on the classifier for that topic class.\footnote{\url{https://github.com/HappyGu0524/MultiControl/blob/main/classify/eval.py}} This would not work in our case (we don't know the topic in advance); instead, we compute the probabilities assigned by each binary classifier to its topic, and then select the topic that has the highest probability. 

\textbf{Keyword control evaluation}. We decided how to compute word coverage based on previous work. We compute Keyword CE as the average over four Word Inclusion Coverage scores, two based on matching words as they appear (Exact Match), and two matching lemmas of words (Lemmatised Match). Following \cite{liu-etal-2023-bolt}, Exact Match considers exact matches between the desired keywords and the words in the generated text, calculated as \textit{Any} (at least one keyword is present), and \textit{All} (all keywords are present). Lemmatised Match, derived from Non-Residual Prompting \cite{carlsson-etal-2022-fine} using the CommonGen code \cite{lin-etal-2020-commongen}, considers lemmas of each word calculated using Spacy (\textit{Cov}) or extending the possible lemmas using Lemmingflect library (\textit{ExtCov}). 

\textbf{Multiple-attribute control evaluation}. In previous work, Multiple-attribute CE is typically calculated as the average of the single-attribute CEs (e.g., sentiment and topic). However, this approach does not provide a true assessment of models in multiple-attribute control tasks, as it does not take into account the requirement that all control attributes must be satisfied \textit{simultaneously}. To address this, we calculate Multiple-attribute CE as the proportion of all single attributes being correct at the same time (i.e.\ \textit{both} sentiment and topic must be correct for an output to be counted as correct overall). To achieve this, we use the single-attribute classifiers described earlier and obtain a final prediction for each attribute by applying a majority vote across the three classifiers for each control attribute. Based on these predictions, we calculate Multiple-attribute CE as the proportion of outputs where all desired attributes are satisfied at the same time. Additionally, we report both the average CE (\textit{Avg} in tables) and the accuracy of both desired sentiment and desired topic at the same time (\textit{Both}) to highlight how much the results can vary depending on the implementation approach.

\subsection{Datasets}
\label{sec:data}

To mitigate the issues related to dataset bias, we prioritise diversity in dataset selection. By using a varied set of datasets that reflect different text generation scenarios, we aim to reduce bias and ensure a more balanced evaluation across systems.

We selected four datasets to test CTG techniques on. Two of these are commonly used free-text generation datasets: the prompts used in the evaluation of
PPLM \cite{dathathri2019plug} which we call PPLM Prompts for short, and the OpenWebText neutral sentiment prompts \cite{Gokaslan2019OpenWeb}. 

We add two datasets used for Story Generation, namely the Cloze Winter 2018 test set \cite{sharma-etal-2018-tackling} and the STS benchmark test set \cite{cer-etal-2017-semeval}. We used only the `main captions' subset of the STS benchmark test set as these are general and can be used to produce texts with all of our control attributes. We use each dataset item to generate one text for each control attribute value, e.g.\ positive and negative for sentiment control, as described in the next section.
Table~\ref{tab:dataset} shows the number of samples contained in each dataset.

\begin{table}[ht]
    \centering
    \small
    \setlength\tabcolsep{2pt} 
    \renewcommand{\arraystretch}{1.10}
    \begin{tabular}{|l|c|}
       \hline
       \textbf{Dataset} & \textbf{\# Samples} \\
       \hline
       PPLM Prompts & 35 \\
       OWT neutral prompts & 5000 \\
       Cloze 2018 test & 1571 \\
       STS benchmark test & 625 \\
       \hline
    \end{tabular}
    \caption{Number of samples for each dataset used for the execution of the experiments.}
    \label{tab:dataset}
\end{table}

\subsection{Obtaining system outputs}\label{sec:obtaining-outputs}

We ensure consistency in the process of running code, generating outputs, and postprocessing results to make sure that all techniques are evaluated under comparable and fair conditions.

Each of our 12 selected CTG techniques is executed using the same datasets and random seeds to ensure comparability and reproducibility. We treat each technique's code as a black box to generate text.

To ensure comparability across systems, we minimally adapt the input text to respect the formatting requirements of each CTG technique, following the reference papers.

For each CTG technique, we minimally adapt the input to respect the formatting required. Input formats vary across techniques: some require only a raw input text (e.g.\ a prompt sentence), while others require both the input text and explicit control attribute value (e.g.\ target sentiment, keywords, or topic). In prompt-based systems, input text and control attribute values are embedded directly within the prompt, whereas other techniques use structured input fields or special tokens. Table~\ref{tab:input_format} provides a detailed overview of the input format used for each CTG technique, specifying whether it includes only the base text, the control attribute values, or a fully formatted prompt that integrates both.

\begin{table}[]
    \centering
    \small
    \setlength\tabcolsep{2pt} 
    \renewcommand{\arraystretch}{1.10}
    \begin{tabular}{|l|l|}
    \hline
        \textbf{Technique} & \textbf{Input format} \\
        \hline
        CTRL & \{control\_attribute\_value\} \{text\}\\
        C BART & \{text\} \{keywords separated by white space\} \\
        Multi CTG & \{text\} \\
        Prior CTG & \{text\} \\
        CAT-PAW & 50256 \{encoded text\} \\
        PPLM & \{control\_attribute\_value\}:\{text\} \\
        Falcon 40B Instruct & \{prompt\} containing \{text\} and \{control\_attribute\_value\} \\
        LLaMa2 70B chat & \{prompt\} containing \{text\} and \{control\_attribute\_value\} \\
        GeDi & \{text\} \\
        DisCup & \{text\} \\
        DExperts & \{text\} \\
        BOLT & \{text\} \\
        \hline
    \end{tabular}
    \caption{Input format for each CTG technique.}
    \label{tab:input_format}
\end{table}

We use the same hyperparameters as provided in the original papers (Appendix~\ref{sec:app_hyperparams}), and the same library versions.

As explained in Section~\ref{sec:sel-methods}, we consider four control attributes in our experiments, namely Sentiment, Topic, Keywords and Multiple-attribute. For each sample in each dataset, we generate one text for each control value associated to the specified control attribute. Table~\ref{tab:control_values} shows the used control values for each control attribute. We use the same keyword sets as Liu et al. \cite{liu-etal-2023-bolt}.

\begin{table}[ht!]
    \centering
    \small
    \setlength\tabcolsep{2pt} 
    \renewcommand{\arraystretch}{1.15}
    \begin{tabular}{|l|p{0.88\textwidth}|}
    \hline
        \textbf{Attribute} & \textbf{Control Values} \\
        \hline
        Sentiment & Positive, Negative\\
        \hline
        Topic & World, Sports, Business, Science/Technology \\
        \hline
        Keywords & [router, Linux, keyboard, server], [plea, subpoena, transcript, bankrupt], [torpedo, headquarters, infantry, battlefield], [court, culture, communism, capitalism], [Bible, church, priest, saint], [microscope, mass, mineral, scientist], [meteor, planet, satellite, astronaut] \\
        \hline
        Multiple & [Positive World], [Positive Sports], [Positive Business], [Positive Science/Technology], [Negative World], [Negative Sports], [Negative Business], [Negative Science/Technology] \\
        \hline
    \end{tabular}
    \caption{Control attributes and respective values used in the experiments.}
    \label{tab:control_values}
\end{table}

In topic control, techniques often use different sets of topic labels, making direct comparison difficult. To address this, we standardise the topic values by identifying a set of topics that were shared across the majority of the selected techniques. We then create a mapping from each label in the standardised set to the system’s original topic labels (see Table~\ref{tab:topic_mapping}). This allows us to ensure that all systems are evaluated on a consistent set of control values for topic generation.

\begin{table}[h!]
    \centering
    \small
    \setlength\tabcolsep{2pt} 
    \renewcommand{\arraystretch}{1.15}
    \begin{tabular}{|l|l|}
        \hline
        \textbf{CTG Technique} & \textbf{Topic Mapping} \\
        \hline
        \multirow{4}{*}{CTRL} & Sports $\rightarrow$ Fitness; \\
         & Business $\rightarrow$ Finance, Legal; \\ 
         & Science/Technology $\rightarrow$ Computing, Science, Technology; \\ 
         & World $\rightarrow$ - \\
        \hline
        \multirow{4}{*}{Prior CTG} & Sports $\rightarrow$ Sports; \\ 
         & Business $\rightarrow$ Business; \\ 
         & Science/Technology $\rightarrow$ Science/Technology; \\ 
         & World $\rightarrow$ World \\
        \hline
        \multirow{4}{*}{CAT-PAW} & Sports $\rightarrow$ -; \\ 
         & Business $\rightarrow$ -; \\ 
         & Science/Technology $\rightarrow$ Science, Computers, Space; \\ 
         & World $\rightarrow$ - \\
        \hline
        \multirow{4}{*}{PPLM} & Sports $\rightarrow$ -; \\ 
         & Business $\rightarrow$ -; \\ 
         & Science/Technology $\rightarrow$ Science, Computers, Space; \\ 
         & World $\rightarrow$ - \\
        \hline
        \multirow{4}{*}{Falcon 40B Instruct} & Sports $\rightarrow$ Sports; \\ 
         & Business $\rightarrow$ Business; \\ 
         & Science/Technology $\rightarrow$ Science/Technology; \\ 
         & World $\rightarrow$ World \\
        \hline
        \multirow{4}{*}{LLaMa2 70B chat} & Sports $\rightarrow$ Sports; \\ 
         & Business $\rightarrow$ Business; \\ 
         & Science/Technology $\rightarrow$ Science/Technology; \\ 
         & World $\rightarrow$ World \\
        \hline
        \multirow{4}{*}{GeDi} & Sports $\rightarrow$ Sports; \\ 
         & Business $\rightarrow$ Business; \\ 
         & Science/Technology $\rightarrow$ Science/Technology; \\ 
         & World $\rightarrow$ World \\
        \hline
    \end{tabular}
    \caption{Mapping to translate the used topic values into topic values available in each CTG technique. Our Topic value $\rightarrow$ Technique Topic value}
    \label{tab:topic_mapping}
\end{table}

Lastly, we postprocess all outputs to extract the generated text, discarding non-text output such as control codes given in the input. Table~\ref{tab:postprocessing} shows the postprocessing applied to generated texts for each CTG technique.

\begin{table}[h!]
    \centering
    \small
    \setlength\tabcolsep{2pt} 
    \renewcommand{\arraystretch}{1.10}
    \begin{tabular}{|l|l|}
    \hline
        \textbf{CTG Technique} & \textbf{Postprocessing} \\
        \hline
        CTRL & remove initial control attribute value \\
        C BART & use text as it is \\
        Multi CTG & use text as it is \\
        Prior CTG & use text as it is \\
        CAT-PAW & remove initial \textbar\textless endoftext\textgreater\textbar token \\
        PPLM & remove initial \textbar\textless endoftext\textgreater\textbar token and control attribute value \\
        Falcon 40B Instruct & remove input prompt and use text between first occurence of "Falcon:" and "User:" \\
        LLaMa2 70B chat & remove input prompt \\
        GeDi & keep text up to and excluding \textbar\textless endoftext\textgreater\textbar token \\
        DisCup & use text as it is \\
        DExperts & use text as it is \\
        BOLT & use text as it is \\
        \hline
    \end{tabular}
    \caption{Postprocessing for each CTG technique.}
    \label{tab:postprocessing}
\end{table}

\section{Controllable Text Generation Techniques Evaluated}
\label{sec:models}

\begin{figure}[t!]
    \centering
    \includegraphics[width=0.48\textwidth]{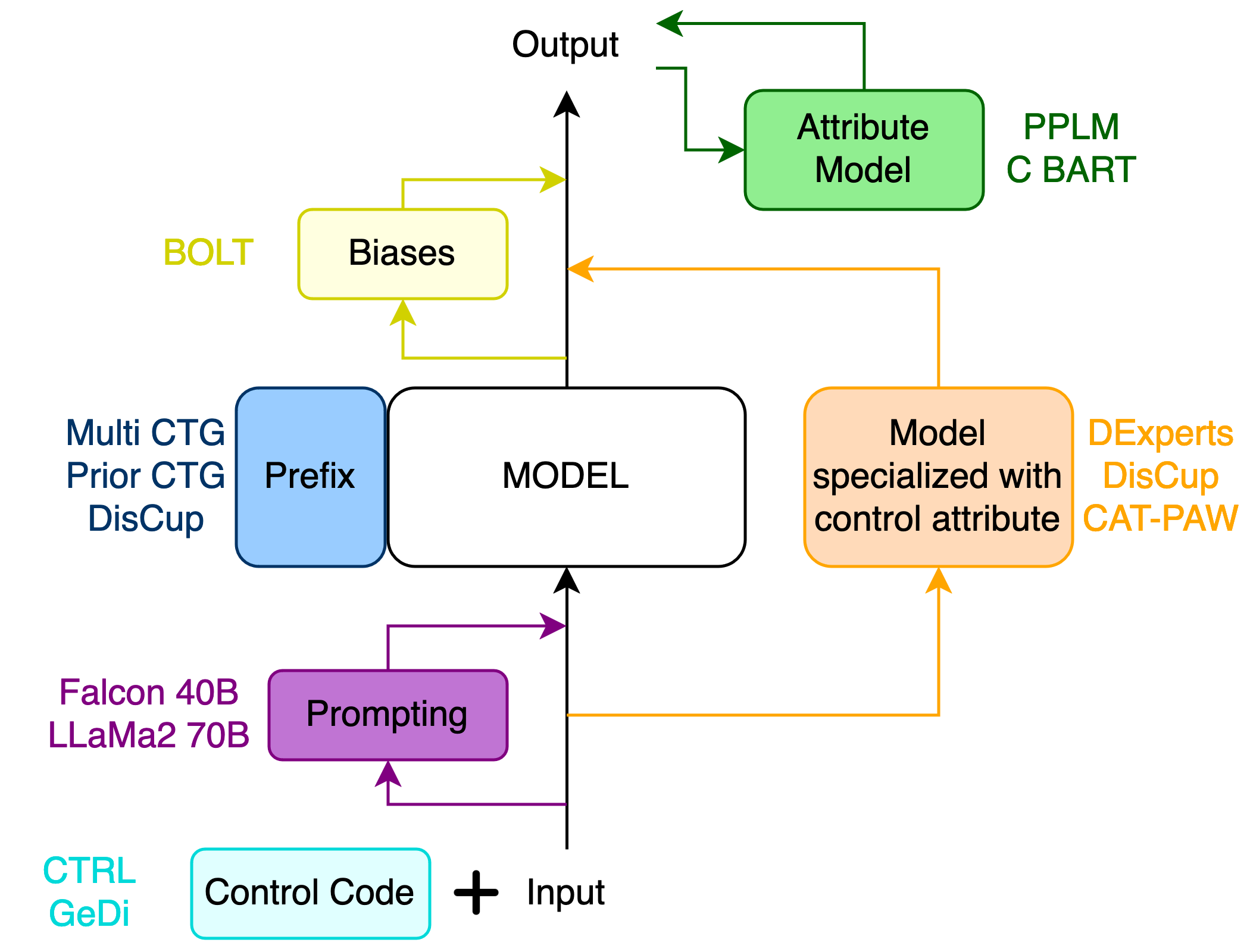}
    \caption{Diagrammatic view of where in the model architecture the control is injected at inference time, for each of our 12 CTG techniques evaluated.}
    \label{fig:control_general}
\end{figure}

In this section, we use the categorisation scheme proposed by Lorandi and Belz \cite{lorandi-belz-2023-control} to characterise how control is implemented in the 12 different techniques we selected for inclusion in our study. Figure~\ref{fig:control_general} provides a diagrammatic view of where in the model architecture the control is injected at inference time.

Below we briefly describe the techniques, grouped together according to control implementation technique as per Lorandi and Belz \cite{lorandi-belz-2023-control}.

\paragraph{Complete Training Techniques.}

CTRL \cite{keskar2019ctrl} is a conditional transformer LM trained to learn control codes which are prepended to the input to steer text generation towards the desired attribute.

\paragraph{Model Fine-Tuning Techniques.}

Controllable BART (C BART) \cite{he-2021-parallel} analyses the input with a token-level classifier to extract information used by the decoder to revise the text by inserting and replacing  tokens with the aim of making sure the sentence contains the desired keywords, while remaining fluent.

Multi CTG \cite{gu-etal-2022-distributional} fine-tunes GPT-2 to learn distinct vector representations for each control attribute. When multiple attributes are desired, the model combines these attribute vectors into a single unified representation, which is then used as a prefix to guide the text generation process toward satisfying all the desired attributes simultaneously.

Applying a similar concept, Prior CTG \cite{gu-etal-2023-controllable} uses prefix-tuning to learn latent representations and applies Normalizing Flow to learn an invertible function to map the latent space into a prior space that is easier to negotiate.

\paragraph{Modification of Token Distribution Techniques.}

PPLM \cite{dathathri2019plug} guides text generation by using pre-trained attribute classifiers without needing to fine-tune the language model itself. During generation, it computes gradients from these external classifiers to determine how each potential next token would affect the desired attributes. These gradients are then used to adjust the language model's predictions, shifting them toward tokens that better align with the desired attributes while maintaining fluency.

ControllAble Text generation with Position-Aware Weighted decoding (CAT-PAW) \cite{gu-etal-2022-improving} introduces a lightweight adjustment mechanism that modifies the language model's output probabilities. This mechanism can either strengthen or weaken the influence of attribute-specific distributions (like sentiment or topic distributions) on the final text generation, by comparing and adjusting the difference between the pre-trained language model's token predictions and the desired attribute-specific token distributions.

\paragraph{Prompting Techniques.}

Falcon 40B Instruct \cite{falcon40B} is a causal decoder-only LM trained on the RefinedWeb dataset \cite{refinedweb}. It adapts the GPT-3 architecture with three main differences: (i) rotational positional embeddings; (ii) multi-query attention, and (iii) parallel attention decoder blocks. 

LLaMa2 70B chat \cite{touvron2023llama} is a transformer-based autoregressive causal LM built on the LLaMa2 model. It is fine-tuned in two steps: (i) supervised fine-tuning using publicly available instruction tuning data, and (ii) Reinforcement Learning with Human Feedback (RLHF), applying two separately optimised reward models.

For both LLMs, we evaluate two prompting strategies: \textit{Zero Shot}, which includes only a simple description of the task, and \textit{Few Shot}, which builds on the task description by adding two example inputs and outputs illustrating the desired control behaviour. Full prompts text used are provided in Appendix~\ref{app:prompts}.

\paragraph{Hybrid Techniques.}
This section's techniques combine multiple control implementation types, e.g.\ Modification of Token Distribution and Model Fine-Tuning. For example, Generative Discriminator (GeDi) \cite{krause-etal-2021-gedi-generative} fine-tunes an LM fine-tunes a language model with control codes that represent different attributes. During generation, GeDi computes the probability distribution for both the desired control code and its opposite, then uses these contrastive signals to guide text generation. By promoting tokens that align with the desired attribute while penalizing those associated with the opposite attribute, GeDi steers the generation toward the target characteristic.

Similarly, Decoding-time Experts (DExperts) \cite{liu-etal-2021-dexperts} guides text generation by using two specialized language models: an expert model fine-tuned on text with desired attributes (e.g., positive sentiment) and an anti-expert model fine-tuned on text with undesired attributes (e.g., negative sentiment). During generation, DExperts combines the predictions from both models with the base language model by boosting token probabilities favored by the expert while suppressing those favored by the anti-expert, allowing for more nuanced control over generated text.

Discriminator Cooperative Unlikelihood Prompt-tuning (DisCup) \cite{zhang-song-2022-discup} fine-tunes an LM incorporating a control attribute discriminator, which evaluates whether generated text exhibits the desired attribute, to optimise control prompts. The optimised prompts are used as prefixes to steer a fixed conditional LM to generate attribute-specific texts. 

Biases Over LogiTs (BOLT) \cite{liu-etal-2023-bolt} steers text generation by adding learnable bias vectors to the language model's logits. These biases are optimized through an energy function that measures both attribute relevance and language fluency: it rewards token choices that align with the desired attribute while penalizing those that harm the text's fluency. During generation, BOLT applies these optimized biases to adjust the model's token probabilities, making attribute-aligned tokens more likely to be selected.

\section{Experimental Set-up}
\label{sec:exp-setup}

We carried out the same experiments for each of the four control attributes from Section~\ref{sec:sel-methods}, Sentiment, Topic, Keywords, and Multiple-attribute, the latter in our study meaning control over both Sentiment and Topic at the same time. All models were executed with their default hyperparameters as presented in the reference papers.  For the complete set of hyperparameters for each model, see Appendix~\ref{sec:app_hyperparams}.

Our experimental grid consisted of combining the 12 techniques from Section~\ref{sec:models} with each of four control attributes,  four datasets, three seeds, and two prompt types in the case of LLMs. Due to execution time (more than 10 days required), BOLT (keywords control) has been executed only with PPLM Prompts (hence the gaps in results bar charts).

\textbf{LLMs}: \{LLaMa2 70B chat, Falcon 40B Instruct\} $\times$ \{zero-shot minimal instruction, few-shot in context\} $\times$ \{Sentiment, Topic, Keywords, Multiple\} $\times$ \{PPLM Prompts, OWT neutral prompts, Cloze dataset, STS benchmark\} $\times$ \{3 seeds (789, 3443, 9817)\}

\textbf{Sentiment}: \{DExperts, GeDi, CAT-PAW, Prior CTG, BOLT, PPLM, DisCup\} $\times$ \{PPLM Prompts, OWT neutral prompts, Cloze dataset, STS benchmark\} $\times$ \{3 seeds (789, 3443, 9817)\}

\textbf{Topic}: \{CTRL, GeDi, CAT-PAW, Prior CTG, PPLM\} $\times$ \{PPLM Prompts, OWT neutral prompts, Cloze dataset, STS benchmark\} $\times$ \{3 seeds (789, 3443, 9817)\}

\textbf{Keywords}: \{C BART, BOLT\} $\times$ \{PPLM Prompts, OWT neutral prompts, Cloze dataset, STS benchmark\} $\times$ \{3 seeds (789, 3443, 9817)\}

\textbf{Multiple}: \{Multi CTG, Prior CTG, PPLM\} $\times$ \{PPLM Prompts, OWT neutral prompts, Cloze dataset, STS benchmark\} $\times$ \{3 seeds (789, 3443, 9817)\}

\section{Evaluation Metrics Analysis}
\label{sec:eval_meth_analysis}

In this Section, we analyse the evaluation metrics from Section~\ref{sec:lev_play_method} to assess their reliability, robustness, and potential biases in evaluating CTG systems. Through this analysis, we aim to gain insights into the different metrics' strengths, weaknesses, sensitivity to model-specific variation, and overall suitability for an LPF evaluation.

\subsection{Classifier testing}
\label{sec:app_classif}

We first tested each classifier on benchmark datasets for sentiment and topic classification. Sentiment classifiers were tested on the Yelp polarity dataset \cite{Zhang2015CharacterlevelCN}, topic classifiers on the AGnews dataset \cite{Zhang2015CharacterlevelCN}. Table~\ref{tab:class_benchm} shows sentiment and topic classification results on these two datasets.

\begin{table}[h!]
    \centering
    \small
    \setlength\tabcolsep{2pt} 
    \renewcommand{\arraystretch}{1.10}
    \begin{tabular}{|l|l|l|cccc|}
    \hline
        \textbf{Control} & \textbf{Benchmark Dataset} & \textbf{Classifier} & \textbf{Acc} & \textbf{F1} & \textbf{Prec} & \textbf{Rec} \\
        \hline
        \multirow{3}{*}{Sentiment} & \multirow{3}{*}{Yelp} & DistilBERT fine-tuned on SST-2 & 90.49 & 90.3 & 92.18 & 88.49 \\
         &  & T5 fine-tuned on SST-2 & 95.17 & 95.26 & 93.63 & 96.95 \\
         &  & DeBERTa fine-tuned on Yelp & 85.7 & 83.59 & 98.02 & 72.86 \\
         \hline
        \multirow{3}{*}{Topic} & \multirow{3}{*}{AGNews} & DistilBERT fine-tuned on AGNews & 94.79 & 94.79 & 94.84 & 94.79 \\
         &  & BERT fine-tuned on AGNews & 93.75 & 93.73 & 93.72 & 93.75 \\
         &  & DeBERTa fine-tuned on AGNews & 94.55 & 94.56 & 94.62 & 94.55 \\
    \hline
    \end{tabular}
    \caption{Accuracy, F1, Precision and Recall of different sentiment and topic classifiers on the Yelp polarity and AGnews datasets.}
    \label{tab:class_benchm}
\end{table}

The results show strong performance for most classifiers, with T5 achieving the highest accuracy and F1 scores for sentiment classification (95.17\% and 95.26, respectively). DistilBERT also performed well, with balanced precision and recall, but DeBERTa showed lower recall (72.86\%) despite a high precision of 98.02\%, suggesting it might overly favour correct classification of one class over another.

For topic classification, all classifiers achieved similarly high performance, with accuracy and F1 scores above 93\%. DistilBERT achieved the highest accuracy (94.79\%), closely followed by DeBERTa and then BERT. These results suggest that all three topic classifiers are reliable for the task, with slight variation in their precision and recall balances.

\subsection{Average Control Effectiveness vs.\ Majority Voting Control Effectiveness}
\label{sec:avg_vs_vote}
In this section, we examine differences between control effectiveness (CE) calculated as the average CE of the single classifier accuracies, vs.\ CE calculated as the majority vote of all classifiers. Tables~\ref{tab:avg_vs_mv_sent}~and~\ref{tab:avg_vs_mv_top} compare results for Sentiment and Topic control, respectively.

\begin{table}[h]
    \begin{subtable}[h]{0.48\textwidth}
        \centering
    \small
    \setlength\tabcolsep{2pt} 
    \renewcommand{\arraystretch}{1.10}
    \begin{tabular}{|ll|cccc|}
        \hline
        \multicolumn{2}{|c|}{\multirow{2}{*}{\textbf{CTG Technique}}} & \multicolumn{4}{c|}{\textbf{Control Effectiveness}$\uparrow$} \\ \cline{3-6}
        & & \multicolumn{2}{c}{\textbf{Average}} & \multicolumn{2}{c|}{\textbf{Majority Vote}} \\
        \hline
        \multicolumn{2}{|l|}{Prior CTG} & \textbf{92.63} (0.5) & [1] & 92.39 (0.5) & [3] \\
        \hline
        \multicolumn{2}{|l|}{CAT PAW} & 61.08 (1.6) & [12] & 60.87 (1.8) & [12] \\
         \multicolumn{2}{|l|}{PPLM} & 73.79 (1.2) & [8] & 73.16 (1.1) & [8] \\
         \hline
        \multirow{3}{*}{Falcon 40B} & ZS & 79.77 (1.9) & [7] & 77.04 (1.8) & [7] \\
         & FS & 65.79 (10.2) & [10] & 66.35 (10.0) & [10] \\
         & Overall & 72.78 (10.1) & [9] & 71.69 (9.0) & [9] \\
        \multirow{3}{*}{LLaMa2 70B} & ZS & 90.74 (2.9) & [4] & 89.1 (0.7) & [4] \\
         & FS & 82.13 (8.7) & [6] & 81.9 (7.3) & [6] \\
         & Overall & 86.43 (7.8) & [5] & 85.5 (6.3) & [5] \\
         \hline
         \multicolumn{2}{|l|}{GeDi} & 91.22 (7.3) & [3] & \textbf{97.4} (0.4) & [1] \\
        \multicolumn{2}{|l|}{DisCup} & 91.52 (2.1) & [2] & 92.9 (1.1) & [2] \\
        \multicolumn{2}{|l|}{DExperts} & 50.41 (0.8) & [13] & 50.34 (1.0) & [13] \\
         \multicolumn{2}{|l|}{BOLT} & 63.61 (1.8) & [11] & 62.9 (1.0) & [11] \\
        \hline
\end{tabular}
       \caption{Sentiment control.}
       \label{tab:avg_vs_mv_sent}
    \end{subtable}
    \hfill
    \begin{subtable}[h]{0.48\textwidth}
        \centering
    \small
    \setlength\tabcolsep{2pt} 
    \renewcommand{\arraystretch}{1.10}
    \begin{tabular}{|ll|cccc|}
        \hline
        \multicolumn{2}{|c|}{\multirow{2}{*}{\textbf{CTG Technique}}} & \multicolumn{4}{c|}{\textbf{Control Effectiveness}$\uparrow$} \\ \cline{3-6}
        & & \multicolumn{2}{c}{\textbf{Average}} & \multicolumn{2}{c|}{\textbf{Majority Vote}} \\
        \hline
        \multicolumn{2}{|l|}{CTRL} & 40.57 (3.1) & [10] & 37.68 (2.7) & [11] \\
        \hline
        \multicolumn{2}{|l|}{Prior CTG} & \textbf{84.28} (3.1) & [1] & 82.69 (0.6) & [2] \\
        \hline
        \multicolumn{2}{|l|}{CAT PAW} & 62.02 (8.7) & [4] & 55.22 (5.4) & [5] \\
        \multicolumn{2}{|l|}{PPLM} & 72.14 (9.7) & [3] & 65.25 (8.5) & [3] \\
        \hline
        \multirow{3}{*}{Falcon 40B} & ZS & 45.87 (5.3) & [8] & 46.84 (1.5) & [8] \\
        & FS & 38.98 (3.7) & [11] & 39.4 (2.0) & [10] \\
        & Overall & 42.43 (5.7) & [9] & 43.12 (4.1) & [9] \\
        \multirow{3}{*}{LLaMa2 70B} & ZS & 55.61 (6.3) & [5] & 55.97 (2.6) & [4] \\
        & FS & 51.12 (10.6) & [7] & 50.7 (10.4) & [7] \\
        & Overall & 53.36 (9.0) & [6] & 53.33 (8.0) & [6] \\
        \hline
        \multicolumn{2}{|l|}{GeDi} & 84.0 (2.9) & [2] & \textbf{84.66} (1.8) & [1] \\
        \hline
\end{tabular}
    \caption{Topic control.}
    \label{tab:avg_vs_mv_top}
     \end{subtable}
     \caption{Comparison between Control Effectiveness computed as Average of the used classifiers and Majority Voting between the classifiers for Sentiment (Pearson’s r = 0.99 with Spearman's $\rho$ 0.98) and Topic (Pearson’s r = 0.98 with Spearman's $\rho$ 0.97). Standard deviation in round brackets. System rank in square brackets.}
     \label{tab:avg_vs_mv}
\end{table}

From these results we can see that the effectiveness of control can vary considerably depending on whether it is computed as the classifier average vs.\ the majority vote. For instance, in sentiment control (Table~\ref{tab:avg_vs_mv_sent}), majority voting results in notably higher scores for some techniques, like GeDi, while causing scores to drop substantially for others, such as Falcon Zero-Shot. This indicates a mixed and technique-dependent impact.

Looking at system ranks (in square brackets) for sentiment control, the only difference is that GeDi and Prior CTG switch ranks from Averge CE to Majority Vote CE. Pearson’s $r$ is high at 0.99, with Spearman's $\rho$ at 0.98. 

In contrast, the results for topic control (Table~\ref{tab:avg_vs_mv_top}) are more varied. Only 5/11 ranks are identical between the two versions of the CE metric. The discrepancies are primarily due to rank inversion of pairs of techniques: CTRL and Falcon FS swap neighbouring ranks, as do CAT PAW and Falcon ZS, and Prior CTG and GeDI once again. Correspondingly, Pearson's is strong at 0.98 with Spearman's $\rho$ at 0.97. 

\begin{figure}
     \centering
     \begin{subfigure}[b]{0.49\textwidth}
         \centering
         \includegraphics[width=\textwidth]{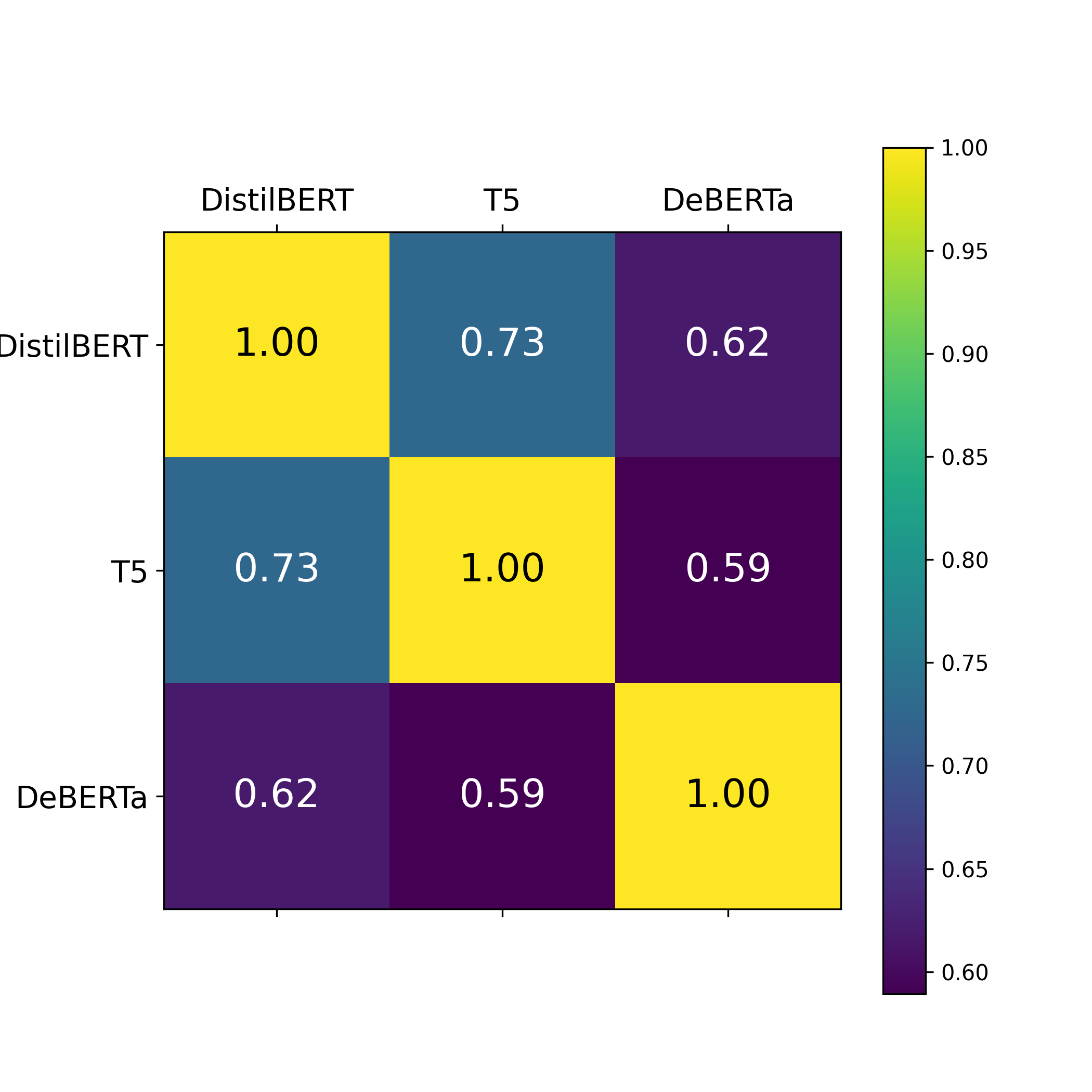}
         \caption{Sentiment control.}
         \label{fig:corr_sent}
     \end{subfigure}
     \hfill
     \begin{subfigure}[b]{0.49\textwidth}
         \centering
         \includegraphics[width=\textwidth]{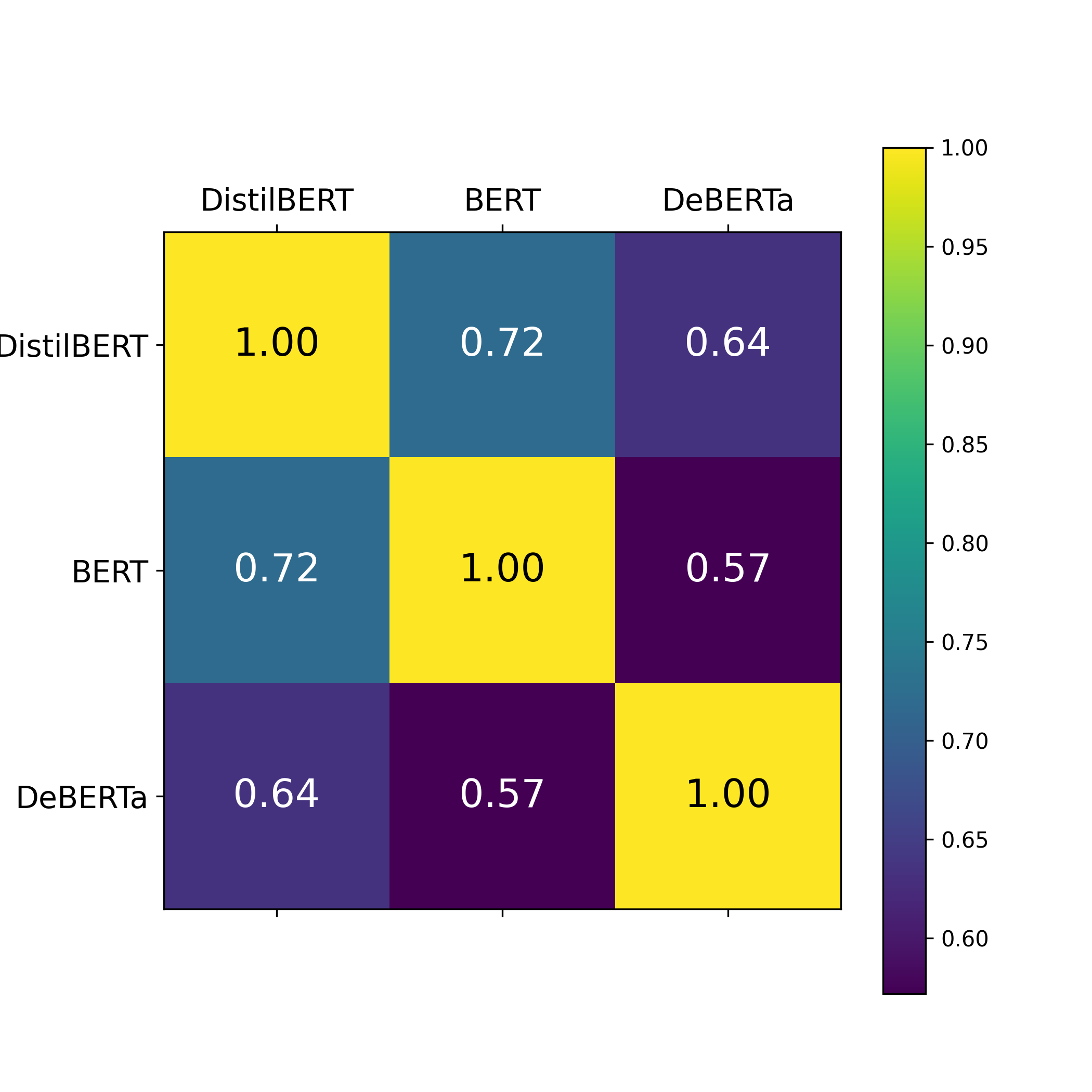}
         \caption{Topic control.}
         \label{fig:corr_topic}
     \end{subfigure}
        \caption{Correlation matrix between classifiers used in sentiment and topic control. Darker color means less correlation.}
        \label{fig:corr}
\end{figure}

To further investigate majority voting control effectiveness and identify any outlier classifiers, we computed the correlations between pairs of classifiers. For both sentiment and topic control (Figures~\ref{fig:corr_sent}~and~\ref{fig:corr_topic} respectively), the DeBERTa classifier shows the least correlation with the other two classifiers. Consequently, majority voting will reflect the views of the other two classifiers to a greater extent, for both control attributes, and is therefore not used in our evaluation. 

\subsection{Perplexity vs.\ SLOR}
\label{sec:ppl_vs_slor}
We also examined the suitability of the SLOR metric vs.\ the more commonly used perplexity measure. \textbf{Perplexity}~\shortcite{jelinek1977perplexity} computes the exponential cross-entropy of a sentence $S$, and captures how likely a PLM is to generate the given text sequence. Assuming that the used PLM captures language knowledge well, then perplexity reflects the quality of the input sequence to some extent \shortcite{wang2022perplexity}, and is often used as a proxy for Fluency. Perplexity is defined as follows:

\begin{equation}
  \label{eq:ppl}
  \text{PPL}(S)=\exp (-\text{NCE}(S))
\end{equation}

\noindent where NCE is the Negative Cross-Entropy which computes the log-probability given an LM, normalised by sentence length:

\begin{equation}
  \label{eq:nce_ppl}
  \text{NCE}(S)=\frac{1}{|S|}\ln (p_{M}(S))
\end{equation}

\noindent where $p_{M}(S)$ is the probability an LM $M$ gives to sentence $S$.

\textbf{SLOR}~\shortcite{kann-etal-2018-sentence} computes the log-probability of a sentence $S$ normalised by unigram log-probability and length:

\begin{equation}
  \label{eq:slor}
  \text{SLOR}(S)=\frac{1}{|S|}(\ln(p_{M}(S))-\ln(p_{u}(S)))
\end{equation}

\noindent where $p_{M}(S)$ is the same as before, while $p_{u}(S)$ is the unigram probability of the sentence:

\begin{equation}
  \label{eq:p_slor}
  p_{u}(S) = \prod_{t\in S}^{} p(t)
\end{equation}

\noindent where $p(t)$ is the probability of token $t$ with no context. SLOR can be seen as NCE normalised by unigram probability.

One limitation of perplexity is that, because it measures the likelihood of the model generating a given text sequence, it may not effectively capture the quality of the text if the PLM considers it unlikely. SLOR was conceived to address this shortcoming.

To compare this and other aspects of the relative performance of the two metrics, we present a side-by-side comparison of evaluation results computed with them, in Tables~\ref{tab:ppl_comp_sent}, \ref{tab:ppl_comp_top}, \ref{tab:ppl_comp_key} and \ref{tab:ppl_comp_multi} for perplexity and SLOR of sentiment, topic, keyword and multiple-attribute control, respectively. We can see, for instance, Falcon and LLaMa2 exhibit very high perplexity across all control attributes (Tables~\ref{tab:ppl_comp_sent}, \ref{tab:ppl_comp_top}, \ref{tab:ppl_comp_key}, and \ref{tab:ppl_comp_multi}). This could be due either to the PLM's inability to recognise high-quality yet less common sequences, or to its failure to accurately model natural language. We can also see from the scores that the SLOR metric fixes this, which it does by accounting for unigram probabilities, thereby mitigating the PLM's bias against less likely but fluent sequences.

In general, we can observe that perplexity and SLOR give very different rankings of the systems. In fact, in sentiment and topic control only one rank remains the same, while in keywords and multiple-attribute control two ranks are the same. Note that the best system identified is consistent between the two metrics. Correspondingly, for sentiment control, overall Pearson's is -0.07 with Spearman's $\rho$ 0.39, but looking at Table~\ref{tab:ppl_comp_sent}, we can observe that scores for LLMs behave in a different way compared to the other techniques. In fact, if we don't include LLMs, Pearson's is a correlation of -0.98 (strongly negative because lower perplexity is better, but higher SLOR is better) with Spearman's 0.97.

A similar picture emerges for topic, keyword, and multiple-attribute control. When correlations are computed across all systems, we observe that Pearson’s $r$ values are -0.38, 0.17, and -0.65, respectively, for topic, keyword, and multiple-attribute control. The corresponding Spearman’s $\rho$ values are 0.69, 0.09, and 0.70. When LLMs are excluded, the correlations strengthen considerably (Pearson’s $r$: -0.86, -1, -0.87; Spearman’s $\rho$: 0.89, 1, 0.99), reinforcing the observation that LLMs behave differently from other CTG techniques.

\begin{table}[ht!]
    \begin{subtable}[h]{0.48\textwidth}
        \centering
    \small
    \setlength\tabcolsep{2pt} 
    \renewcommand{\arraystretch}{1.10}
    \begin{tabular}{|ll|cc|cc|}
        \hline
        \multicolumn{2}{|c|}{\textbf{Technique}} & \multicolumn{2}{c|}{\textbf{Perplexity}$\downarrow$} & \multicolumn{2}{c|}{\textbf{SLOR}$\uparrow$} \\
        \hline
        \multicolumn{2}{|l|}{Prior CTG} & 74.55 (4.8) & [5] & 12.01 (0.1) & [8] \\
        \hline
        \multicolumn{2}{|l|}{CAT PAW} & 128.75 (15.0) & [8] & 10.69 (0.1) & [13] \\
        \multicolumn{2}{|l|}{PPLM} & 29.91 (3.1) & [2] & 12.9 (0.1) & [3] \\
        \hline
        \multirow{3}{*}{Falcon 40B} & ZS & 1910.86 (688.4) & [11] & 12.66 (0.1) & [5] \\
        & FS & 4326.4 (641.4) & [13] & 11.65 (0.6) & [10] \\
        & Ov & 3118.63 (1378.9) & [12] & 12.15 (0.7) & [7] \\
        \multirow{3}{*}{LLaMa2 70B} & ZS & 46.82 (27.2) & [3] & 12.93 (1.0) & [2] \\
         & FS & 1481.46 (819.5) & [10] & 12.63 (0.3) & [6] \\
        & Ov & 764.14 (922.4) & [9] & 12.78 (0.7) & [4] \\
        \hline
        \multicolumn{2}{|l|}{GeDi} & 98.91 (8.9) & [6] & 11.55 (0.1) & [11] \\
        \multicolumn{2}{|l|}{DisCup} & 117.52 (17.3) & [7] & 11.43 (0.2) & [12] \\
        \multicolumn{2}{|l|}{DExperts} & 64.91 (10.5) & [4] & 11.92 (0.2) & [9] \\
        \multicolumn{2}{|l|}{BOLT} & \textbf{17.0} (1.4) & [1] & \textbf{13.23} (0.1) & [1] \\
        \hline
    \end{tabular}
       \caption{Sentiment control.}
       \label{tab:ppl_comp_sent}
    \end{subtable}
    \hfill
    \begin{subtable}[h]{0.48\textwidth}
    \centering
    \small
    \setlength\tabcolsep{2pt} 
    \renewcommand{\arraystretch}{1.10}
    \begin{tabular}{|ll|cc|cc|}
        \hline
        \multicolumn{2}{|c|}{\textbf{Technique}} & \multicolumn{2}{c|}{\textbf{Perplexity}$\downarrow$} & \multicolumn{2}{c|}{\textbf{SLOR}$\uparrow$} \\
        \hline
        \multicolumn{2}{|l|}{CTRL} & \textbf{17.87} (6.6) & [1] & \textbf{14.98} (0.3) & [1] \\
        \hline
        \multicolumn{2}{|l|}{Prior CTG} & 152.47 (12.1) & [7] & 11.15 (0.1) & [11] \\
        \hline
        \multicolumn{2}{|l|}{CAT PAW} &19.53 (1.4) & [2] & 12.96 (0.1) & [4] \\
        \multicolumn{2}{|l|}{PPLM} & 28.04 (3.9) & [4] & 13.07 (0.1) & [3] \\
        \hline
        \multirow{3}{*}{Falcon 40B} & ZS & 1677.81 (487.3) & [9] & 12.82 (0.1) & [6] \\
        & FS & 4878.04 (909.6) & [11] & 11.58 (0.7) & [9] \\
        & Ov & 3277.93 (1758.6) & [10] & 12.2 (0.8) & [8] \\
        \multirow{3}{*}{LLaMa2 70B} & ZS & 24.76 (11.3) & [3] & 13.4 (0.0) & [2] \\
        & FS & 216.04 (273.9) & [8] & 12.51 (0.4) & [7] \\
        & Ov & 120.4 (216.1) & [6] & 12.96 (0.5) & [4] \\
        \hline
        \multicolumn{2}{|l|}{GeDi} & 112.39 (8.5) & [5] & 11.51 (0.1) & [10] \\
        \hline
    \end{tabular}
        \caption{Topic control.}
        \label{tab:ppl_comp_top}
     \end{subtable}
    
    \begin{subtable}[h]{0.48\textwidth}
        \centering
    \small
    \setlength\tabcolsep{2pt} 
    \renewcommand{\arraystretch}{1.10}
    \begin{tabular}{|ll|cc|cc|}
        \hline
        \multicolumn{2}{|c|}{\textbf{Technique}} & \multicolumn{2}{c|}{\textbf{Perplexity}$\downarrow$} & \multicolumn{2}{c|}{\textbf{SLOR}$\uparrow$} \\
        \hline
        \multicolumn{2}{|l|}{CTG BART} & 694.23 (127.5) & [2] & 9.75 (0.1) & [7] \\
        \hline
        \multirow{3}{*}{Falcon 40B} & ZS & 7486.81 (614.9) & [8] & 11.6 (0.0) & [5] \\
        & FS & 3180.63 (769.6) & [5] & 11.82 (0.6) & [3] \\
        & Ov & 5333.72 (2263.0) & [7] & 11.71 (0.4) & [4] \\
        \multirow{3}{*}{LLaMa2 70B} & ZS & \textbf{455.21} (433.4) & [1] & \textbf{13.17} (0.1) & [1] \\
        & FS & 3242.02 (4012.5) & [6] & 11.29 (1.5) & [6] \\
        & Ov & 1848.61 (3175.8) & [4] & 12.23 (1.4) & [2] \\
        \hline
        \multicolumn{2}{|l|}{BOLT} & 1298.33 (76.9) & [3] & 9.07 (0.1) & [8] \\
        \hline
    \end{tabular}
       \caption{Keywords control.}
       \label{tab:ppl_comp_key}
    \end{subtable}
    \hfill
    \begin{subtable}[h]{0.48\textwidth}
        \centering
    \small
    \setlength\tabcolsep{2pt} 
    \renewcommand{\arraystretch}{1.10}
    \begin{tabular}{|ll|cc|cc|cc|}
        \hline
        \multicolumn{2}{|c|}{\textbf{Technique}} & \multicolumn{2}{c|}{\textbf{Perplexity}$\downarrow$} & \multicolumn{2}{c|}{\textbf{SLOR}$\uparrow$} \\
        \hline
        \multicolumn{2}{|l|}{Multi CTG} & 95.73 (11.4) & [4] & 11.44 (0.3) & [8] \\
        \multicolumn{2}{|l|}{Prior CTG} & 46.65 (1.6) & [3] & 11.88 (0.6) & [5] \\
        \hline
        \multicolumn{2}{|l|}{PPLM} & 35.58 (3.2) & [2] & 12.66 (0.1) & [3] \\
        \hline
        \multirow{3}{*}{Falcon 40B} & ZS & 12328.26 (853.0) & [9] & 11.25 (0.1) & [9] \\
        & FS & 3418.53 (468.3) & [7] & 11.82 (0.6) & [6] \\
        & Ov & 7873.4 (4507.7) & [8] & 11.54 (0.5) & [7] \\
        \multirow{3}{*}{LLaMa2 70B} & ZS & \textbf{33.58} (23.0) & [1] & \textbf{13.11} (0.6) & [1] \\
        & FS & 492.76 (139.6) & [6] & 12.28 (0.8) & [4] \\
        & Ov & 263.17 (250.4) & [5] & 12.7 (0.8) & [2] \\
        \hline
    \end{tabular}
        \caption{Multiple-attribute control.}
        \label{tab:ppl_comp_multi}
     \end{subtable}
     \caption{Comparison between Perplexity and SLOR for Sentiment, Topic, Keywords, and Multiple-attribute control. Standard deviation in round brackets. System rank in square brackets. ZS=Zero Shot, FS=Few Shot, Ov=Overall.}
     \label{tab:ppl_comp}
\end{table}

\section{Results}
\label{sec:results}

Tables~\ref{tab:sent_res_overall}, \ref{tab:top_res_overall}, \ref{tab:key_res_overall} and \ref{tab:multi_res_overall} show per-system Distinct-n (diversity), SLOR (fluency), and CE (control effectiveness) results as weighted averages over all datasets and seeds considering the datasets' size as weights, 
for Sentiment, Topic, Keywords and Multiple-attribute control.

Figures~\ref{fig:sentiment_bars}, \ref{fig:topic_bars}, \ref{fig:keywords_bars} and \ref{fig:multiple_bars} provide a closer look at control effectiveness, separating out results for the different control attribute value and dataset combinations, for Sentiment, Topic, Keywords and Multiple-attribute control, in each of the figures respectively. In the following subsections, we discuss both types of results for each control type in turn.

\subsection{Sentiment Control}

Looking at the Sentiment Control results in Table~\ref{tab:sent_res_overall}, in terms of Distinct-n, we can see that the hybrid techniques outperform all the other techniques with DExperts and GeDi having the best overall performance for $n=1$ and $n=2, 3$, respectively. In terms of SLOR, the highest score is achieved by BOLT, but the other techniques have broadly similar scores. 

Control Effectiveness results differ for different classifiers, showing the unreliability of using a single classifier as evaluation metric. E.g.\ GeDi outperforms Prior CTG according to the DistilBERT and T5 classifiers, but DeBERTa ranks them the other way round. Prior CTG is among the techniques evaluated consistently across all classifiers, and has the highest average Control Effectiveness. Furthermore, for both Falcon and LLaMa2 few-shot performs worse than zero-shot, showing that the addition of examples does not help the task.

\begin{figure*}[ht!]
    \centering
    \includegraphics[width=0.99\textwidth]{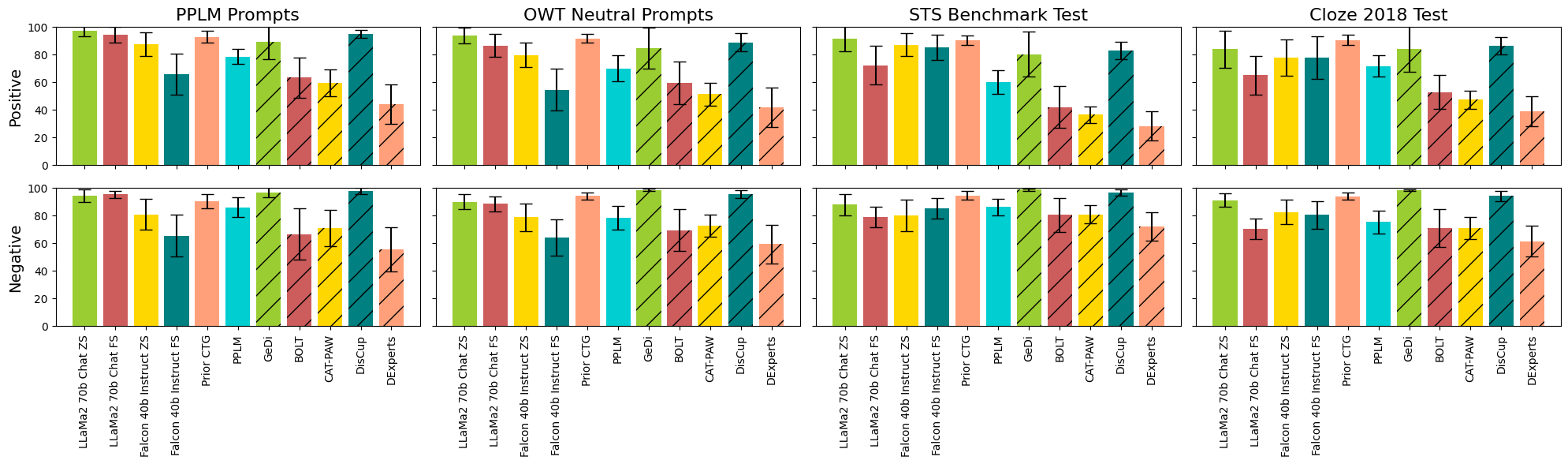}
    \caption{Evaluation results for Sentiment control showing the average system-level Average Control Effectiveness for each combination of dataset and control attribute values (positive/negative). Standard deviation depicted as error bars.}
    \label{fig:sentiment_bars}
\end{figure*}

Looking at Control Effectiveness for different individual control values and dataset combinations  (Figure~\ref{fig:sentiment_bars}), we can see that Prior CTG has consistently high results. GeDi achieves almost perfect results for Negative sentiment with all datasets, but results are less good for Positive where it also has large stdev. DisCup is strong on negative too, with less of a drop off and lower stdev for positive. In general, we see a difference between datasets, e.g.\ when tested on PPLM Prompts all models achieve higher overall performance.

\begin{table*}[h!]
    \centering
    \small
    \setlength\tabcolsep{2pt} 
    \renewcommand{\arraystretch}{1.15}
    \begin{tabular}{|ll|ccc|c|cccc|}
\hline
\multicolumn{2}{|c|}{\multirow{2}{*}{\textbf{CTG Technique}}} & \multicolumn{3}{c|}{\textbf{Distinct-n}$\uparrow$} & \multirow{2}{*}{\textbf{SLOR}$\uparrow$} & \multicolumn{4}{c|}{\textbf{Control Effectiveness} $\uparrow$} \\ \cline{3-5} \cline{7-10}
& & \textbf{dist1} & \textbf{dist2} & \textbf{dist3} & & \textbf{DistilBERT} & \textbf{T5} & \textbf{DeBERTa} & \textbf{Avg} \\
 \hline
\multicolumn{10}{|c|}{\textit{Model Fine-Tuning}}\\
\hline
\multicolumn{2}{|l|}{Prior CTG} & 7.68 (3.3) & 38.77 (6.2) & 69.14 (4.8) & 12.01 (0.1) & 92.77 (0.4) & 92.82 (0.4) & \textbf{92.3} (0.4) & \textbf{92.63} (0.5) \\
\hline
\multicolumn{10}{|c|}{\textit{Modification of Token Distribution}} \\
\hline
\multicolumn{2}{|l|}{CAT PAW} & 5.6 (1.7) & 33.8 (3.1) & 65.3 (2.6) & 10.7 (0.1) & 61.2 (1.5) & 60.8 (1.9) & 61.2 (1.1) & 61.1 (1.6) \\
\multicolumn{2}{|l|}{PPLM} & 5.7 (2.1) & 32.7 (3.5) & 62.3 (2.5) & 12.9 (0.1) & 74.8 (0.8) & 73.3 (1) & 73.3 (1.0) & 73.8 (1.2) \\
\hline
\multicolumn{10}{|c|}{\textit{Prompting}} \\
\hline
\multirow{3}{*}{Falcon 40B Instruct} & ZS & 5.1 (2.7) & 33.2 (5) & 63.3 (4.1) & 12.7 (0.1) & 79.8 (1.4) & 79.7 (1.8) & 79.9 (2.2) & 79.8 (1.9) \\
& FS & 8.3 (2.8) & 42.1 (3.7) & 67.9 (3.4) & 11.6 (0.6) & 66.3 (10.7) & 67.2 (10.1) & 63.8 (9.4) & 65.8 (10.2) \\
& Ov & 6.7 (3.2) & 37.7 (6.3) & 65.6 (4.4) & 12.1 (0.7) & 73 (10.2) & 73.4 (9.6) & 71.9 (10.6) & 72.8 (10.1) \\
\cline{2-10}
\multirow{3}{*}{LLaMa2 70B Chat} & ZS & 5.2 (2.7) & 32.3 (5.0) & 61.1 (3.9) & 12.9 (1.0) & 90.6 (1.1) & 92.3 (0.5) & 89.3 (4.3) & 90.7 (2.9) \\
& FS & 8.7 (4.6) & 41.8 (6.4) & 68 (3.1) & 12.6 (0.3) & 82.4 (7.8) & 84.6 (7.7) & 79.3 (9.5) & 82.1 (8.7) \\
& Ov & 6.9 (4.2) & 37 (7.5) & 64.6 (4.9) & 12.8 (0.7) & 86.5 (6.9) & 88.4 (6.7) & 84.3 (8.9) & 86.4 (7.8) \\
\hline
\multicolumn{10}{|c|}{\textit{Hybrid}} \\
\hline
\multicolumn{2}{|l|}{GeDi} & 10.1 (5.1) & \textbf{62.8} (4.7) & \textbf{88.1} (1.2) & 11.5 (0.1) & \textbf{95} (0.4) & \textbf{97.6} (0.7) & 81 (1.1) & 91.2 (7.3) \\
\multicolumn{2}{|l|}{DisCup} & 9.5 (3.3) & 52.8 (4.0) & 82.5 (0.7) & 11.4 (0.2) & 92.9 (0.7) & 92.5 (1.7) & 89.1 (1.1) & 91.5 (2.1) \\
\multicolumn{2}{|l|}{DExperts} & \textbf{13.2} (3.8) & 56.1 (3.0) & 80.8 (0.8) & 11.9 (0.2) & 50.5 (0.9) & 50.3 (0.9) & 50.4 (0.5) & 50.4 (0.8) \\
\multicolumn{2}{|l|}{BOLT} & 7.9 (3.1) & 43 (3.9) & 72.8 (2.0) & \textbf{13.2} (0.1) & 65 (1.7) & 62.6 (1.4) & 63.2 (1.5) & 63.6 (1.8) \\
\hline
\end{tabular}
\caption{Evaluation results for Sentiment control. Each row contains the weighted average over all datasets and 3 seeds. The Control Effectiveness columns show the accuracy of each classifier separately and the average between them. Best score per column in bold. Standard deviation in round brackets. ZS=Zero-Shot, FS=Few-Shot, Ov=Overall results averaged across ZS and FS.
}
\label{tab:sent_res_overall}
\end{table*}

\subsection{Topic control}

Table~\ref{tab:top_res_overall} shows the results of Topic control. In terms of Distinct-n, CTRL displays unusually low scores compared to other techniques, indicating limited lexical diversity likely due to its strong reliance on control codes. While this behaviour is expected given CTRL's design, it contrasts with its high SLOR score, suggesting that despite low diversity, its outputs remain relatively fluent and well-formed. In fact, CTRL has a higher SLOR score than the other techniques by some margin.

In general, results are lower for topic control than for sentiment control in terms of Control Effectiveness with the same techniques (Table~\ref{tab:top_res_overall}). In terms of Average CE, Prior CTG and GeDi lead by a hefty margin. While PPLM outperforms Prior CTG according to the DeBERTa classifier, Prior CTG outperforms all other techniques according to the DistilBERT and BERT classifiers and the average score. As before, there are differences between results depending on which classifier is used, highlighting once again the need to rely on multiple classifiers for a fairer evaluation. Similar to sentiment control, LLaMa2 has higher results when using Zero-Shot prompts.

\begin{figure*}[h!]
    \centering
    \includegraphics[width=0.99\textwidth]{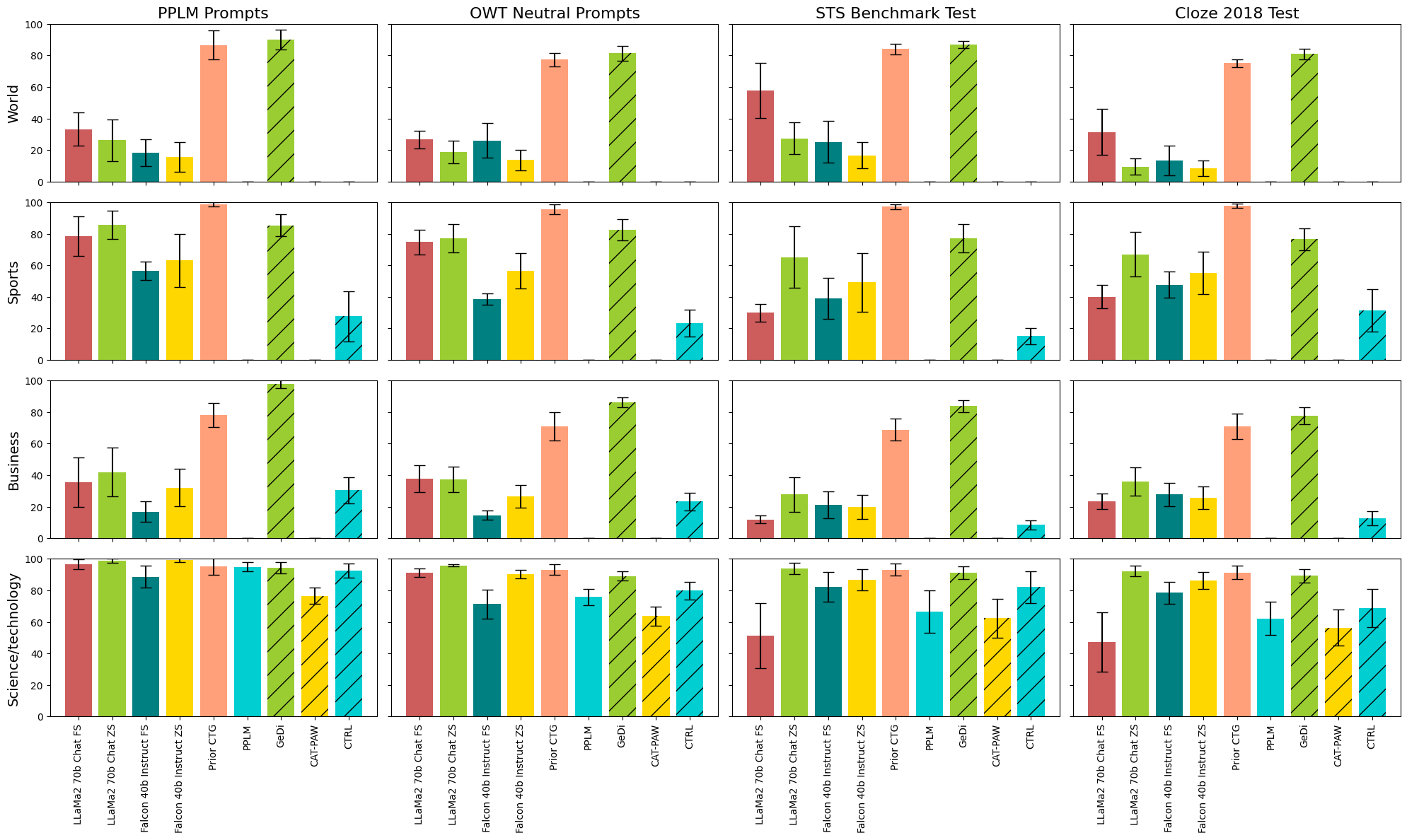}
    \caption{Evaluation results for Topic control showing the average Control Effectiveness across the 4 used datasets and control attribute values. Standard deviation depicted as error bars.}
    \label{fig:topic_bars}
\end{figure*}

Analysing the differences between control values (Figure~\ref{fig:topic_bars}), we notice that results for Science/Technology are higher across all models than the other control values. On the other hand, results for Business and World are lowest, except for Prior CTG and GeDi which are able to achieve good results in all control values across all datasets.

\begin{table*}
    \centering
    \small
    \setlength\tabcolsep{2pt} 
    \renewcommand{\arraystretch}{1.15}
    \begin{tabular}{|ll|ccc|c|ccc|c|}
\hline
\multicolumn{2}{|c|}{\multirow{2}{*}{\textbf{CTG Technique}}} & \multicolumn{3}{c|}{\textbf{Distinct-n}$\uparrow$} & \multirow{2}{*}{\textbf{SLOR}$\uparrow$} & \multicolumn{4}{c|}{\textbf{Control Effectiveness} $\uparrow$} \\ \cline{3-5} \cline{7-10}
& & \textbf{dist1} & \textbf{dist2} & \textbf{dist3} & & \textbf{DistilBERT} & \textbf{BERT} & \textbf{DeBERTa} & \textbf{Avg} \\
\hline
\multicolumn{10}{|c|}{\textit{Complete Training}}\\
\hline
\multicolumn{2}{|l|}{CTRL} & 0.6 (0.4) & 4.2 (0.6) & 8.5 (0.9) & \textbf{15} (0.3) & 42.2 (4.0) & 38.9 (2.2) & 40.6 (1.7) & 40.6 (3.1) \\
\hline
\multicolumn{10}{|c|}{\textit{Model Fine-Tuning}}\\
\hline
\multicolumn{2}{|l|}{Prior CTG} & \textbf{9.6} (4.4) & 47.9 (6.1) & 76.5 (3.2) & 11.1 (0.1) & \textbf{87} (0.7) & \textbf{85.6} (1.0) & 80.2 (1.3) & \textbf{84.3} (3.1) \\
\hline
\multicolumn{10}{|c|}{\textit{Modification of Token Distribution}}\\
\hline
\multicolumn{2}{|l|}{CAT PAW} & 6.8 (2.6) & 38.9 (3.6) & 70.3 (1.6) & 13 (0.1) & 58.3 (5.8) & 55.2 (4.2) & 72.6 (2.2) & 62 (8.7) \\
\multicolumn{2}{|l|}{PPLM} & 6.0 (2.4) & 36.2 (3.4) & 66.9 (2.0) & 13.1 (0.1) & 68.6 (8.7) & 65.9 (7.3) & \textbf{81.9} (2.9) & 72.1 (9.7) \\
\hline
\multicolumn{10}{|c|}{\textit{Prompting}}\\
\hline
\multirow{3}{*}{Falcon 40B Instruct} & ZS & 5.1 (2.6) & 33.3 (5.5) & 64.3 (4.6) & 12.8 (0.1) & 50.3 (1.5) & 48.5 (1.4) & 38.8 (2.0) & 45.9 (5.3) \\
& FS & 8.3 (2.7) & 42.8 (3.8) & 69.4 (2.8) & 11.6 (0.7) & 41 (2.6) & 40.8 (3.2) & 35.1 (1.0) & 39 (3.7) \\
& Ov & 6.7 (3.1) & 38.1 (6.7) & 66.8 (4.6) & 12.2 (0.8) & 45.7 (5.1) & 44.6 (4.6) & 37 (2.4) & 42.4 (5.7) \\
\cline{2-10}
\multirow{3}{*}{LLaMa2 70B Chat} & ZS & 5.2 (2.8) & 31.5 (5.6) & 59.4 (4.8) & 13.4 (0) & 59.2 (2.5) & 59.9 (2.3) & 47.6 (3.7) & 55.6 (6.3) \\
& FS & 9.2 (5.0) & 41.3 (6.8) & 66.3 (3.3) & 12.5 (0.4) & 53.7 (10.6) & 53.5 (11.5) & 46.2 (7.7) & 51.1 (10.6) \\
& Ov & 7.2 (4.5) & 36.4 (7.9) & 62.8 (5.4) & 13 (0.5) & 56.5 (8.2) & 56.7 (8.9) & 46.9 (6.1) & 53.4 (9.0) \\
\hline
\multicolumn{10}{|c|}{\textit{Hybrid}}\\
\hline
\multicolumn{2}{|l|}{GeDi} & 8.7 (5.1) & \textbf{61.9} (6.0) & \textbf{89.1} (1.7) & 11.5 (0.1) & 85.8 (1.7) & 85.4 (2.0) & 80.7 (1.3) & 84 (2.9) \\
\hline
\end{tabular}
\caption{Results for Topic control. Control Effectiveness shows the accuracy of the 3 
classifiers and the average between them. Numbers are weighted averages over all datasets and 3 seeds. Best score per column in bold. Standard deviation in round brackets. ZS=Zero-Shot, FS=Few-Shot, Ov=Overall results averaged across ZS and FS.
}
\label{tab:top_res_overall}
\end{table*}

\subsection{Keyword control}

Table~\ref{tab:key_res_overall} shows the overall results of Keywords control. Considering Distinct-n, prompting techniques (Falcon and LLaMa2) show relatively low scores, indicating limited lexical diversity compared to fine-tuned or hybrid approaches. In contrast, BOLT outperforms all other techniques by a margin.

SLOR results for Keyword Control (Table~\ref{tab:key_res_overall}) for every technique are worse compared to sentiment and topic control. LLaMa2 achieves the highest SLOR, but is lower than the best SLOR in topic control.

In terms of Control Effectiveness, C BART achieves almost perfect performance for keyword control. All  techniques are good at including at least one exact keyword in the generated text (`Any' column), but performance drops dramatically for correctly inserting all keywords (`All' column). Falcon improves its performance when using Few-Shot prompting in all metrics except Exact Match / All, while LLaMa2 still shows better results when using Zero-Shot prompting in all metrics except distinct-n with n=1,2,3.

\begin{figure*}[h!]
    \centering
    \includegraphics[width=1\textwidth]{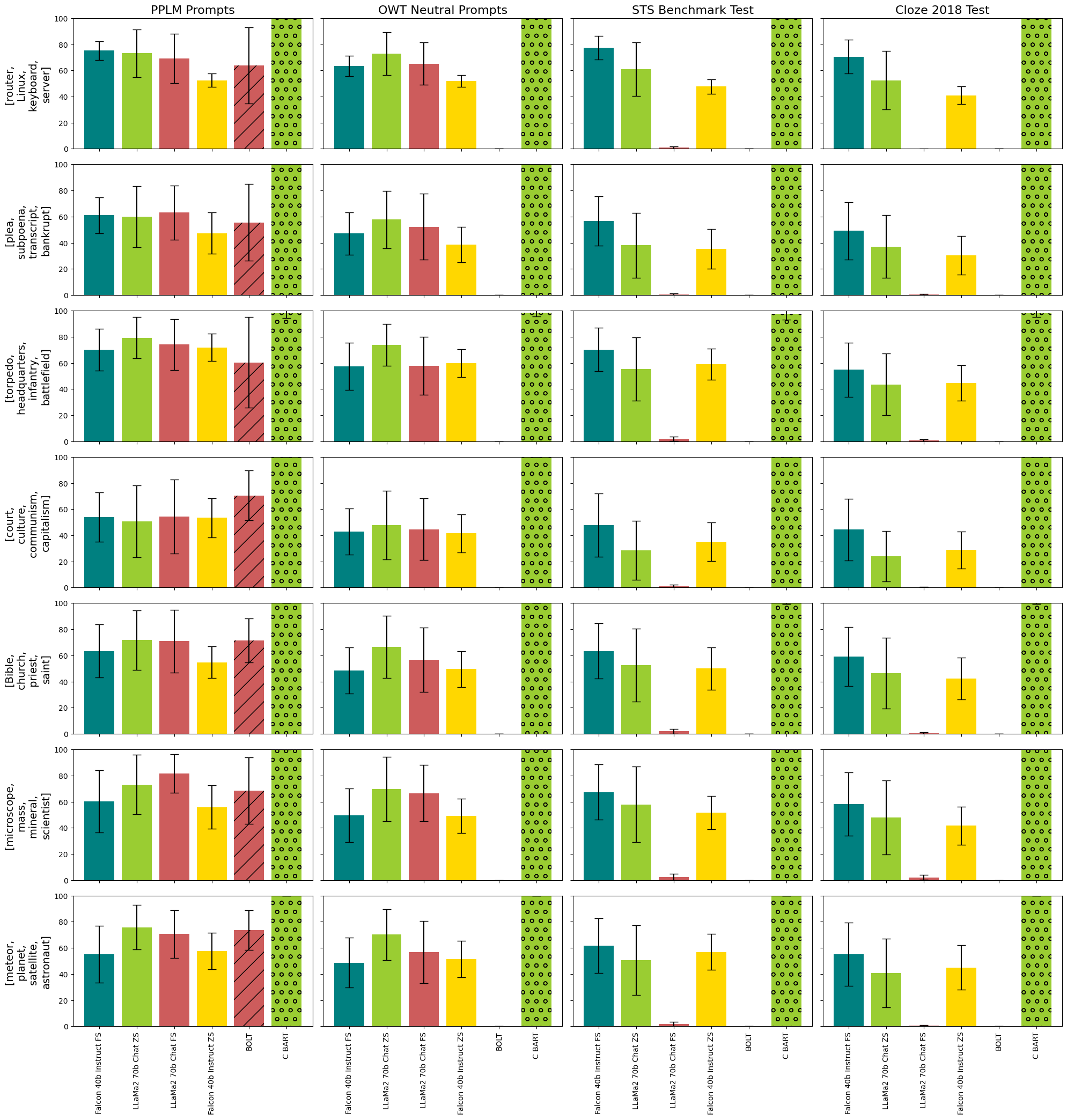}
    \caption{Evaluation results for Keywords control showing the average Control Effectiveness across the 4 used datasets and control attribute values. Standard deviation depicted as error bars.}
    \label{fig:keywords_bars}
\end{figure*}

Looking at differences in Control Effectiveness for different dataset and keyword set combinations (Figure~\ref{fig:keywords_bars}), we can see that for the STS benchmark and Cloze 2018 test sets, LLaMa2's performance collapses completely in the few-shot setting; it is much better preserved in the zero-shot setting. For the other techniques, the performance across different keywords sets are overall stable. The standout performance across all settings is by C BART which succeeds in including all keywords in nearly all cases; even in those cases, the problem is not with C BART itself, but with the evaluation metric (given the perfect exact-match performance in Table~\ref{tab:key_res_overall}, it's likely that the lemmatiser made mistakes).

\begin{table*}[t!]
    \centering
    \small
    \setlength\tabcolsep{1.8pt} 
    \renewcommand{\arraystretch}{1.10}
\begin{tabular}{|ll|ccc|c|cc|cc|c|}
\hline
\multicolumn{2}{|c|}{\multirow{3}{*}{\textbf{CTG Technique}}} & \multicolumn{3}{c|}{\multirow{2}{*}{\textbf{Distinct-n}$\uparrow$}} & \multirow{3}{*}{\textbf{SLOR}$\uparrow$} & \multicolumn{5}{c|}{\textbf{Control Effectiveness} $\uparrow$} \\ \cline{7-11}
& & & & & & \multicolumn{2}{c|}{\textbf{Exact Match}} & \multicolumn{2}{c|}{\textbf{Lemmatization}} & \multirow{2}{*}{\textbf{Avg}} \\ \cline{3-5} \cline{7-10}
& & \textbf{dist1} & \textbf{dist2} & \textbf{dist3} & & \textbf{Any} & \textbf{All} & \textbf{ExtCov} & \textbf{Cov} & \\
\hline
\multicolumn{11}{|c|}{\textit{Model Fine-Tuning}}\\
\hline
\multicolumn{2}{|l|}{C BART} & 10.1 (1) & 33.1 (1.5) & 45 (1.5) & 9.7 (0.1) & \textbf{100} (0) & \textbf{100} (0) & \textbf{100} (0.0) & \textbf{98.9} (0.2) & \textbf{99.7} (0.5) \\
\hline
\multicolumn{11}{|c|}{\textit{Prompting}}\\
\hline
\multirow{3}{*}{Falcon 40B Instruct} & ZS & 5.2 (2.7) & 30.8 (5.9) & 58.7 (5.5) & 11.6 (0) & 63.6 (2.7) & 28.8 (4.8) & 48.4 (4.4) & 46.1 (4.3) & 46.7 (13.0) \\
& FS & 6.3 (2.3) & 34.2 (3.6) & 61.8 (3.1) & 11.8 (0.6) & 77.1 (5.5) & 27.0 (2.9) & 55.6 (4.1) & 53.3 (4.4) & 53.2 (18.3) \\
& Ov & 5.7 (2.6) & 32.5 (5.2) & 60.2 (4.7) & 11.7 (0.4) & 70.3 (8) & 27.9 (4.1) & 52 (5.6) & 49.7 (5.6) & 50 (16.2) \\ \cline{2-11}
\multirow{3}{*}{LLaMa2 70B Chat} & ZS & 4.4 (2.6) & 25.8 (6.2) & 49.4 (7.2) & \textbf{13.2} (0.1) & 89.8 (6.7) & 27.6 (11.0) & 60.2 (11.5) & 57.9 (11.2) & 58.9 (24.3) \\
& FS & 10.9 (7.2) & 37.5 (10.3) & 54.8 (3.7) & 11.3 (1.5) & 62.5 (39.7) & 17.7 (11.7) & 40.9 (26.5) & 39.6 (25.7) & 40.1 (31.9) \\
& Ov & 7.7 (6.3) & 31.6 (10.3) & 52.1 (6.3) & 12.2 (1.4) & 76.2 (31.6) & 22.7 (12.4) & 50.5 (22.6) & 48.8 (21.9) & 49.5 (29.9) \\
\hline
\multicolumn{11}{|c|}{\textit{Hybrid}}\\
\hline
\multicolumn{2}{|l|}{BOLT} & \textbf{40.5} (0.4) & \textbf{82.8} (0.4) & \textbf{93.9} (0.3) & 9.1 (0.1) & 99 (0.8) & 26.7 (7.8) & 65.9 (2.6) & 64.7 (2.9) & 64.1 (26) \\
\hline
\end{tabular}
\caption{Evaluation results for Keywords control. Control Effectiveness shows the Word Inclusion Coverage considering exact match and lemmatization. Each row contains the weighted average over all datasets and 3 seeds. Best score per column in bold. Standard deviation in round brackets. Cov=Word Inclusion Coverage with lemmas extracted with Spacy, ExtCov=Word Inclusion Coverage with lemmas extracted with Spacy and Lemmingflect, ZS=Zero-Shot, FS=Few-Shot, Ov=Overall results averaged across ZS and FS.}
\label{tab:key_res_overall}
\end{table*}

\subsection{Multiple-attribute control}

Table~\ref{tab:multi_res_overall} reports results for distinct-n and SLOR as before, but for CE, recall the discussion from Section~\ref{sec:app_classif}: it's the `Both' column that gives a true assessment of effective control over both attributes \textit{simultaneously}. We include separate results for sentiment control and topic control for reference, as well as the average between them for comparison with other research which typically reports multiple-attribute control effectiveness in this way. 

Regarding Distinct-n, prompting techniques achieve lower lexical diversity compared to other techniques, while PPLM achieves the highest scores, reflecting greater output variability. In terms of SLOR, LLaMa2 achieves the highest scores, indicating strong fluency despite its lower diversity.

Overall, performance drops for multiple-attribute control compared to our other control types, as techniques now need to satisfy two control attributes at the same time (Table~\ref{tab:multi_res_overall}). In terms of SLOR, PPLM outperforms the other techniques; Falcon Zero-Shot has the worst score. 

In terms of Control Effectiveness, Falcon and LLaMa2 performance are higher when using Zero-Shot prompting compared to Few-Shot.

PPLM outperforms the other techniques when computing Control Effectiveness in terms of both control attributes being correct at the same time (`Both' column) and the average between sentiment and topic. Overall we see performance drop for Prior CTG which has good performance in controlling single attributes, but not multiple ones.

\begin{figure*}[ht!]
    \centering
    \includegraphics[width=0.92\textwidth]{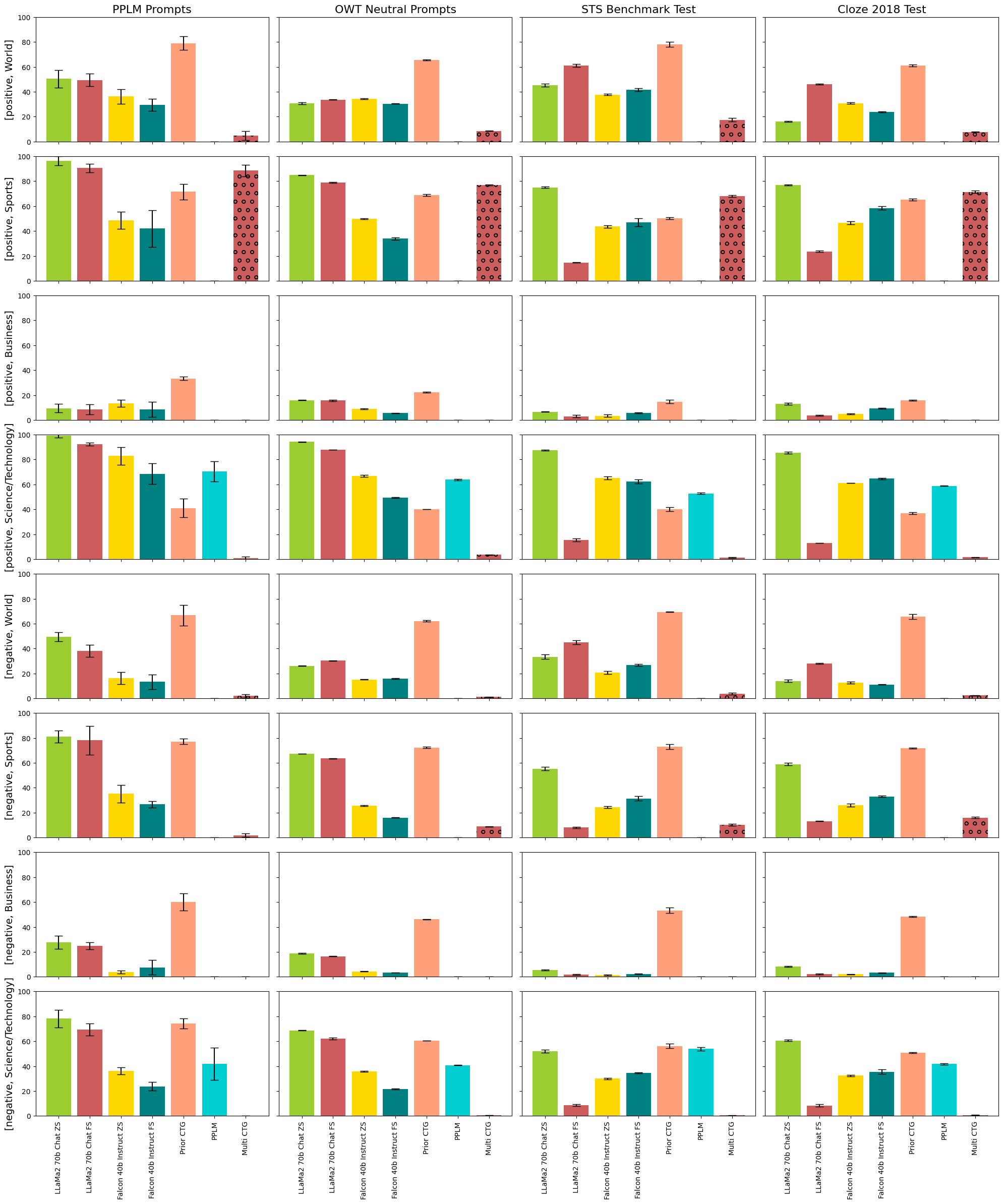}
    \caption{Evaluation results for Multiple-attribute control showing the average Control Effectiveness across the 4 used datasets and control attribute values. Standard deviation reported with error bars.}
    \label{fig:multiple_bars}
\end{figure*}

Looking at the differences between datasets and control values in Figure~\ref{fig:multiple_bars}, all models performs worst at generating texts with positive/negative sentiment combined with the Business topic for all datasets. The highest results are achieved for positive sentiment combined with Sports or Science/Technology. This is in line with what was previously observed in topic control.

\begin{table*}[h!]
    \centering
    \small
    \setlength\tabcolsep{3pt} 
    \renewcommand{\arraystretch}{1.15}
\begin{tabular}{|ll|ccc|c|c|cc|c|}
\hline
\multicolumn{2}{|c|}{\multirow{2}{*}{\textbf{CTG Technique}}} & \multicolumn{3}{c|}{\textbf{Distinct-n}$\uparrow$} & \multirow{2}{*}{\textbf{SLOR}$\uparrow$} & \multicolumn{4}{c|}{\textbf{Control Effectiveness} $\uparrow$} \\ \cline{3-5} \cline{7-10}
& & \textbf{dist1} & \textbf{dist2} & \textbf{dist3} & & \textbf{Both} & \textbf{S} & \textbf{T} & \textbf{Avg}\textsuperscript{(!)} \\
\hline
\multicolumn{10}{|c|}{\textit{Model Fine-Tuning}}\\
\hline
\multicolumn{2}{|l|}{Prior CTG} & 3.5 (2.1) & 17.9 (4.8) & 36.3 (5.1) & 12.4 (0) & \textbf{53.9} (1.5) & 85.5 (0.5) & 64.9 (1.5) & \textbf{75.2} (0.9) \\
\multicolumn{2}{|l|}{Multi CTG} & 0.5 (1.3) & 6.3 (2.6) & 13.8 (3.3) & 11.4 (0.3) & 12.5 (0.1) & 50.1 (0.3) & 24.9 (0.1) & 37.5 (0.2) \\
\hline
\multicolumn{10}{|c|}{\textit{Modification of Token Distribution}}\\
\hline
\multicolumn{2}{|l|}{PPLM} & \textbf{5.7} (2) & \textbf{32.5} (3.7) & \textbf{61.5} (3.1) & 12.7 (0.1) & 51.9 (1.3) & 59.7 (1.5) & \textbf{86.6} (2.5) & 73.2 (1) \\
\hline
\multicolumn{10}{|c|}{\textit{Prompting}}\\
\hline
\multirow{3}{*}{Falcon 40B Instruct} & ZS & 2.3 (1.9) & 24.2 (5) & 54.7 (4.7) & 11.2 (0.1) & 29.3 (1.3) & 67.9 (0.7) & 40.7 (2.1) & 54.3 (0.9) \\
& FS & 3.1 (2) & 27.8 (3.8) & 55.4 (3.2) & 11.8 (0.6) & 24.6 (3.9) & 63.7 (7.9) & 37.9 (1.4) & 50.8 (4.6) \\
& Ov & 2.8 (2) & 26 (4.8) & 55 (4) & 11.5 (0.5) & 26.9 (3.7) & 65.8 (6) & 39.3 (2.2) & 52.5 (3.7) \\ \cline{2-10}
\multirow{3}{*}{LLaMa2 70B Chat} & ZS & 1.6 (1.7) & 18.3 (4.9) & 43.3 (5.5) & \textbf{13.1} (0.6) & 48.4 (4) & \textbf{87.6} (1.4) & 55.7 (3.9) & 71.6 (2.5) \\
& FS & 3.7 (3.1) & 24.2 (4.7) & 46.3 (4.2) & 12.3 (0.8) & 39.3 (14.1) & 77.5 (11.9) & 48.7 (12.9) & 63.1 (12.4) \\
& Ov & 2.6 (2.7) & 21.2 (5.6) & 44.8 (5.1) & 12.7 (0.8) & 43.9 (11.3) & 82.6 (9.8) & 52.2 (10.1) & 67.4 (9.9) \\
\hline
\end{tabular}
\caption{Results for Multiple-attribute control. Control Effectiveness is calculated via majority voting between three classifiers. The `Both' column shows the proportion of outputs where both control attributes are correct. The \textbf{Avg CE}\textsuperscript{(!)} column shows the average of the single-attribute CE scores and is included for reference only. Each row contains the weighted average over all datasets and 3 seeds. Best score per column in bold. Standard deviation in round brackets. CI=Control Implementation, MFT=Model Fine-Tuning, MTD=Modification of Token Distribution, P=Prompting, 
S=Sentiment, T=Topic, ZS=Zero-Shot, FS=Few-Shot, Ov=Overall results averaged across ZS and FS.}
\label{tab:multi_res_overall}
\end{table*}

\section{Efficiency Assessment}
\label{sec:eff_res}
We further analyse the different CTG techniques in terms of memory space required to store them and inference execution time. Assessing efficiency is crucial, because (i) performance needs to always be seen relative to cost, and (ii) the scalability and practicality of deploying techniques in real-world applications depends directly on efficiency, particularly for resource-constrained environments or large-scale systems.

Table~\ref{tab:mem_gb} shows the memory usage (in GB) of each CTG technique during inference. Some CTG techniques are model-agnostic and can, in principle, be applied to any pretrained LM. As a result, the memory usage of such techniques can vary significantly depending on the size of the underlying LM. Among the techniques evaluated, the LLMs \textit{Prompting} approaches use by far the largest amount of memory. In contrast, techniques based on \textit{Model Fine-Tuning}, \textit{Modification of Token Distribution}, and \textit{Hybrid} techniques use the least memory space while also achieving the highest performance in sentiment, topic, and keywords control.

\begin{table}[ht!]
    \centering
    \small
    \setlength\tabcolsep{3pt} 
    \renewcommand{\arraystretch}{1.10}
    \begin{tabular}{|l|l|c|}
        \hline
        \multirow{2}{*}{\textbf{CTG Technique}} & \multicolumn{2}{c|}{\textbf{Memory used (GB)}} \\ \cline{2-3}
        & \textbf{Details} & \textbf{Total} \\
        \hline
        \multicolumn{3}{|c|}{\textit{Complete Training}} \\
        \hline
        CTRL & CTRL model ($\sim$14.4) & $\sim$14.4 \\
        \hline
        \multicolumn{3}{|c|}{\textit{Model Fine-Tuning}} \\
        \hline
        C BART & GPT-2 ($\sim$0.5) + C BART large ($\sim$1.6) & $\sim$2.1 \\
        Multi CTG & BERT base ($\sim$0.4) + GPT-2 medium ($\sim$1.5) + Multi model ($\sim$4.6) & $\sim$6.5 \\
        Prior CTG & BERT base ($\sim$0.4) + GPT-2 medium ($\sim$1.5) + Prior model ($\sim$4.7) & $\sim$6.6 \\
        \hline
        \multicolumn{3}{|c|}{\textit{Modification of Token Distribution}} \\
        \hline
        CAT PAW & 2* GPT-2 medium ($\sim$1.5) & $\sim$3 \\
        PPLM & GPT-2 medium ($\sim$1.5) & $\sim$1.5 \\
        \hline
        \multicolumn{3}{|c|}{\textit{Prompting}} \\
        \hline
        Falcon 40B Instruct & Falcon model ($\sim$83.6) & $\sim$83.6 \\
        LLaMa2 70B Chat & LLaMa model ($\sim$137.9) & $\sim$137.9 \\
        \hline
        \multicolumn{3}{|c|}{\textit{Hybrid}} \\
        \hline
        GeDi & GPT-2 xl ($\sim$6.4) + Sentiment model ($\sim$1.5) + Topic model ($\sim$1.5) & $\sim$9.4 \\
        DisCup & GPT-2 large ($\sim$3.2) + Positive Sentiment ($\sim$0.9) + Negative Sentiment ($\sim$0.9) & $\sim$5 \\
        DExperts & GPT-2 large ($\sim$3.2) + Positive Sentiment ($\sim$3) + Negative Sentiment ($\sim$3) & $\sim$9.2 \\
        BOLT & GPT-2 large ($\sim$3.2) + Sentiment model ($\sim$0.5) & $\sim$3.7 \\
        \hline
    \end{tabular}
    \caption{Memory used to store each CTG technique in GB.}
    \label{tab:mem_gb}
\end{table}

Table~\ref{tab:exec_1_sample} shows the average inference time per sample for each CTG technique for each control attribute explored. To compute this, we recorded the start and end time of each generation and averaged the time taken across individual samples over all datasets and seeds. Cells marked with -- indicate that the technique does not support the corresponding control attribute and was therefore not executed for that setting. Among the evaluated approaches, \textit{Modification of Token Distribution} techniques (Table~\ref{tab:exec_1_sample}) require more inference time compared to other technique types. These methods often need to iterate over the sentence to dynamically adjust the token distribution to match the desired attribute, adding computational complexity.

\begin{table}[ht!]
    \centering
    \small
    \setlength\tabcolsep{3pt} 
    \renewcommand{\arraystretch}{1.10}
    \begin{tabular}{|ll|c|c|c|c|}
        \hline
        \multicolumn{2}{|l}{\textbf{CTG Technique}} & \textbf{Sentiment} & \textbf{Topic} & \textbf{Keywords} & \textbf{Multiple} \\
        \hline
        \multicolumn{6}{|c|}{\textit{Complete Training}} \\
        \hline
        \multicolumn{2}{|l|}{CTRL} & -- & 01.93 $\pm$0.04 & -- & -- \\
        \hline
        \multicolumn{6}{|c|}{\textit{Model Fine-Tuning}} \\
        \hline
        \multicolumn{2}{|l|}{C BART} & -- & -- & \textbf{00.05} $\pm$0.0 & -- \\
        \multicolumn{2}{|l|}{Multi CTG} & -- & -- & -- & 07.92 $\pm$0.01 \\
        \multicolumn{2}{|l|}{Prior CTG} & 00.35 $\pm$0.0 & \textbf{00.31} $\pm$0.0 & -- & 03.46 $\pm$0.03 \\
        \hline
        \multicolumn{6}{|c|}{\textit{Modification of Token Distribution}} \\
        \hline
        \multicolumn{2}{|l|}{CAT PAW} & 14.69 $\pm$1.5 & 12.09 $\pm$0.35 & -- & -- \\
        \multicolumn{2}{|l|}{PPLM} & 37.42 $\pm$0.13 & 11.71 $\pm$0.06 & -- & 14.58 $\pm$0.07 \\
        \hline
        \multicolumn{6}{|c|}{\textit{Prompting}} \\
        \hline
        \multirow{2}{*}{Falcon 40B Instruct} & Zero-Shot & 01.37 $\pm$0.03 & 01.35 $\pm$0.01 & 01.46 $\pm$0.0 & \textbf{01.38} $\pm$0.0 \\
        & Few-Shot & 03.20 $\pm$0.25 & 02.91 $\pm$0.02 & 03.13 $\pm$0.02 & 02.96 $\pm$0.0 \\ \cline{2-6}
        \multirow{2}{*}{LLaMa2 70B chat} & Zero-Shot & 01.91 $\pm$0.0 & 01.92 $\pm$0.0 & 02.01 $\pm$0.0 & 01.93 $\pm$0.01 \\
        & Few-Shot & 02.41 $\pm$0.0 & 02.41 $\pm$0.0 & 02.38 $\pm$0.01 & 02.37 $\pm$0.0 \\
        \hline
        \multicolumn{6}{|c|}{\textit{Hybrid}} \\
        \hline
        \multicolumn{2}{|l|}{GeDi} & 02.84 $\pm$0.04 & 03.15 $\pm$0.01 & -- & -- \\
        \multicolumn{2}{|l|}{DisCup} & 00.5 $\pm$0.01 & -- & -- & -- \\
        \multicolumn{2}{|l|}{DExperts} & \textbf{00.19} $\pm$0.0 & -- & -- & -- \\
        \multicolumn{2}{|l|}{BOLT} & 22.80 $\pm$0.14 & -- & 288.69 $\pm$0.0 & -- \\
        \hline
    \end{tabular}
    \caption{Inference execution time in seconds per one sample for each CTG technique using the different control attributes (sentiment, topic, keywords, and multiple-attribute). Each row contains the average over all datasets and 3 seeds with the standard deviation. In bold the fastest technique per column. Cells marked with `--` indicate that the technique does not support the corresponding control attribute and was therefore not executed for that setting.}
    \label{tab:exec_1_sample}
\end{table}

In contrast, \textit{Complete Training}, \textit{Model Fine-Tuning}, and \textit{Prompting} generally achieve low inference times as they rely on a single model without external intervention. However, few-shot prompting introduces some additional overhead, although it remains competitive.

The performance of \textit{Hybrid} techniques varies as it heavily depends on the specific implementation and the control attributes involved.

Understanding these trade-offs is crucial for selecting the most appropriate technique based on specific use cases and computational constraints.

\section{Comparison with Original Results}
\label{sec:comp_ref_papers}

In order to provide an idea of the difference that LPF principles can make, in Table~\ref{tab:comp_ref_ce} we contrast some of the CE results from our LPF evaluation (averaged over the same three classifiers for all systems, see Section~\ref{sec:lev_play_method} for more details) with those from the papers in which the systems were reported for the first time (a different single classifier in each case). Note that we can only include those systems where we have results from the original paper for a dataset from our LPF set of datasets. We do not have access to all datasets from the original works for two main reasons: (i) some datasets used in the original papers are not publicly available, or the testing splits are not shared; and (ii) the details about dataset usage in some papers are unclear. This limitation reduces the number of systems we can directly compare.

Out of the 12 systems selected for evaluation, we are able to compare 6 systems with their original results. This subset provides a meaningful comparison to highlight the impact of standardised and consistent evaluation conditions on CE performance assessment results.
\begin{table}[]
    \centering
    \small
    \setlength\tabcolsep{2pt} 
    \renewcommand{\arraystretch}{1.10}
\begin{tabular}{|l|l|l|cc|}
\hline
\textbf{Control} & \textbf{Dataset} & \textbf{CTG Technique} & \textbf{LPF} & \textbf{Original} \\
\hline
\multirow{6}{*}{Sentiment} & \multirow{4}{*}{PPLM Prompts} & Prior CTG & 91.59 & 99.7 \\
 & & CAT-PAW & 65.24 & 52.08 \\
 & & BOLT & 64.92 & 80.12 \\
 & & PPLM & 82.22 & 78.8 \\
\cline{2-5}
 & \multirow{2}{*}{OWT neutral prompts} & DExperts & 50.53 & 95.4 \\
 & & DisCup & 92.18 & 94.31 \\
 \hline
 \multirow{2}{*}{Topic} & \multirow{2}{*}{PPLM Prompts} & Prior CTG & 89.68 & 97.8 \\
  & & CAT-PAW & 76.51 & 49.09 \\
 \hline
\end{tabular}
    \caption{Side by side comparison of Control Effectiveness scores for sentiment control and topic control systems as assessed in original work (Original)\ and in LPF where it is the average of three classifier scores (DistilBERT, BERT, DeBERTa for topic; DistilbERT, T5, DeBERTa for sentiment). All scores express the proportion of outputs that are classified as having the property they are intended to have.}
    \label{tab:comp_ref_ce}
\end{table}
Numbers in Table~\ref{tab:comp_ref_ce} are classification accuracies representing the proportion of outputs that are classified as having the intended property, with the controlled property (e.g.\ sentiment = positive) providing the gold labels. E.g.\ row 1 in the table compares the three-classifier LPF CE accuracy (91.59) with the single-classifier accuracy from the original paper (99.7). Most CE accuracies paired in this way are broadly in the same ballpark, but some stand out for particularly large differences (e.g.\ DExperts on OWT Neutral Prompts with 50.53 vs. 95.4 and CAT-PAW on PPLM Prompts with 76.51 49.09). In 5 out of 8 cases, our LPF evaluation results in lower scores than in the original paper, the two CAT-PAW results and the PPLM results are much \textit{better} in our evaluation than the original.

Most CE accuracies paired in this way are broadly in the same ballpark, but some stand out for particularly large differences (e.g.\ DExperts on OWT Neutral Prompts and CAT-PAW on PPLM Prompts). In 5 out of 8 cases, our LPF evaluation results in lower scores than in the original paper, such as for DExperts on OWT Neutral Prompts (50.53 vs. 95.4), BOLT on PPLM Prompts (64.92 vs. 80.12). In contrast, the two CAT-PAW results and the PPLM results are much \textit{better} in our evaluation than the original on both sentiment (65.24 vs. 52.08) and topic (76.51 vs. 49.09) control. Overall, Pearson’s r for LPF and Original scores is just 0.31 with Spearman's $\rho$ 0.37 and is not statistically significant.

\section{Discussion}
\label{sec:disc}

The results presented in this study highlight the advantages of employing an LPF framework for the evaluation of CTG systems. Beyond the specific findings related to CTG, the LPF approach has broader implications for evaluation practices in NLP and beyond.

\textbf{Generalisation beyond CTG.} While this work focuses on CTG systems, the principles of LPF evaluation standardisation, such as fairness, and diversity in evaluation metrics, are applicable to other NLP tasks, such as summarisation, and dialogue generation. These tasks can also suffer from similar issues of inconsistent datasets, metric implementation biases, and ad hoc experimental setups. Extending the LPF approach to other domains could promote more reliable and reproducible comparisons across systems, enabling a better understanding of model strengths and weaknesses in varied contexts.

\textbf{Relation to Benchmarks.} Benchmarks play a crucial role in evaluating NLP systems, but they are not without limitations. While standardised benchmarks provide a shared foundation for comparisons, they often fall short in addressing biases introduced by metric implementation or dataset selection. Furthermore, benchmarks tend to favour static evaluation setups, which may not account for variability in real-world applications or the evolving nature of NLP tasks. By complementing benchmarks with LPF principles, such as multiple metrics and diverse datasets, we can address some of these shortcomings and ensure a more holistic evaluation framework.

\textbf{Challenges of Reporting Results for Multiple Metrics.} Reporting results across a large number of metrics, even within the same evaluation context, can provide a multi-faceted view of system performance, but it also introduces challenges. Different metrics, even when applied to the same outputs, may prioritise different aspects of performance and yield conflicting indications of quality. This can obscure rather than clarify the overall evaluation, making it harder to draw reliable conclusions. Furthermore, reporting too many metrics risks diluting the significance of genuinely informative measures and introducing noise into the evaluation. Careful consideration of which metrics best capture key dimensions of performance is therefore important for maintaining clarity and comparability across systems.

\textbf{Towards a Balanced Evaluation Framework.} The LPF framework, as demonstrated in this study, seeks to balance the need for standardisation with the flexibility to account for variability in evaluation conditions. By employing consistent datasets and evaluation protocols while integrating diverse metrics and classifiers, we have aimed to take a step towards fairer comparison of systems that can be expected to yield truer results. However, future work should explore the scalability of this approach, particularly in tasks involving more complex evaluation setups.

\section{Limitations}
\label{sec:lim}

This paper presents a comparative evaluation of CTG systems using an LPF framework. However, several limitations should be acknowledged. First, we focused on CTG techniques that implement only three control attributes: sentiment, topic, and keywords. While these attributes are widely used and representative of CTG, they do not capture the full range of control attributes explored in the literature, such as style, knowledge, or syntax constraints. Further experiments are needed to extend our analysis to techniques that support additional control attributes. Moreover, the sets of values for each control attribute (e.g., binary sentiment or a fixed set of topics) were limited. Expanding these sets could provide further insights into the generalisability of the evaluated techniques.

Second, the selected CTG techniques were restricted to those published in the ACL Anthology. While this ensures the inclusion of state-of-the-art techniques and reproducibility based on publicly available implementations, it excludes techniques presented in other venues or those without accessible resources. Expanding the scope to include non-ACL Anthology techniques could potentially provide additional insights. 

Third, our evaluation relies exclusively on automatic metrics. Although automatic metrics are essential for large-scale comparisons and reproducibility, they may not fully capture nuanced aspects of text quality, such as coherence, creativity, or appropriateness, which are better assessed through human evaluation. Incorporating human evaluation would provide complementary insights.

\section{Conclusion}
\label{sec:conclusion}

In this paper, we have proposed an approach to evaluating Controlled Text Generation (CTG) techniques which aims to level the evaluation playing field and ensure fair, reliable, and reproducible comparisons. Our method incorporates four key components: (i) systematic selection of CTG techniques to reduce bias, (ii) evaluation on multiple datasets to improve generalisability, (iii) consistent and comparable methods for generating system outputs, and (iv) the use of diverse and fair evaluation metrics to assess system performance. 

We evaluated 12 state-of-the-art CTG techniques with the proposed LPF approach, across four control attributes (sentiment, topic, keywords, multiple-attribute control), and four diverse datasets (PPLM Prompts, OWT neutral prompts, STS benchmark test, Cloze 2018 test). This enabled us to generate outputs and evaluate them under consistent conditions, ensuring that all systems were assessed fairly and comprehensively.

Our results highlight several key findings.
First, specialised CTG techniques consistently outperform general-purpose LLMs, with Fine-Tuning and Hybrid approaches demonstrating superior overall performance across all evaluation metrics. Furthermore, we observed that techniques evaluated under the LPF framework often performed worse than reported in their original papers. This discrepancy highlights the impact of standardised and unbiased evaluation conditions, which eliminate favourable biases inherent in prior comparisons. Additionally, the inclusion of diverse evaluation metrics, such as fluency, diversity, and control effectiveness, provided a more nuanced understanding of the strengths and weaknesses of different techniques.

These findings underscore the critical importance of fairness and systematicity in CTG evaluation. Our LPF framework mitigates biases introduced by inconsistent datasets, metric implementations, and evaluation protocols, offering a more accurate comparison of techniques. Moreover, the insights gained from this study have implications beyond CTG, demonstrating the potential for LPF principles to improve evaluation in other NLP tasks such as summarisation, and dialogue generation.

In conclusion, our proposed approach represents a significant step towards fairer and reproducible NLP system evaluation. By setting a higher standard for fairness, we aim to contribute to the broader adoption of robust evaluation methodologies, and ultimately to accelerating progress in the field.

\begin{acks}
This work was conducted with the financial support of the Research Ireland Centre for Research Training in Digitally-Enhanced Reality (d-real) under Grant No. 18/CRT/6224. 
This work was also conducted with the support of Research Ireland under Grant Agreement No.\ 13/RC/2106\_P2 at the ADAPT SFI Research Centre at Dublin City University. For the purpose of Open Access, the author has applied a CC BY public copyright licence to any Author Accepted Manuscript version arising from this submission.
\end{acks}

\bibliographystyle{ACM-Reference-Format}
\bibliography{sample-base}

\appendix

\section{LLMs Prompts}\label{app:prompts}

Tables~\ref{tab:temp-zero-shot-falc} and~\ref{tab:temp-few-shot-falc} show the zero-shot and few-shot prompts used with Falcon 40B Instruct model, while Tables~\ref{tab:temp-zero-shot-llama} and~\ref{tab:temp-few-shot-llama} show the prompts used with LLaMa 2 70B chat model.

\begin{table*}[h!]
\centering\small
\renewcommand{\arraystretch}{1.15}
\begin{tabular}{|>{\raggedright\arraybackslash}p{0.15\textwidth}|>{\raggedright\arraybackslash}p{0.75\textwidth}|}
\hline
\multicolumn{2}{|c|}{\textbf{Falcon 40B Instruct Zero-Shot prompt}} \\
\hline
\textbf{Template for PPLM Prompts and OWT neutral prompts:} & System: You are a text generator. Reply with only one turn.\par User: Write a fluent English text starting with "\{text\}" with \{control attribute value\} \{control attribute\}\par Falcon: \\
\hline
\textbf{Template for STS benchmark test and Cloze 2018 test:} & System: You are a story writer. Reply with only one turn.\par User: Write a fluent English story starting with "\{text\}" with \{control attribute value\} \{control attribute\}\par Falcon: \\
\hline
\textbf{Multiple-attribute control instruction template:} & User: Write a fluent English \{story or text\} starting with "\{text\}" with \{sentiment value\} sentiment and \{topic value\} topic \\
\hline
\end{tabular}
\caption{\label{tab:temp-zero-shot-falc}
Template of the Zero-Shot prompt for Falcon 40B Instruct model.}
\end{table*}

\begin{table*}[h!]
\centering\small
\renewcommand{\arraystretch}{1.15}
\begin{tabular}{|>{\raggedright\arraybackslash}p{0.15\textwidth}|>{\raggedright\arraybackslash}p{0.75\textwidth}|}
\hline
\multicolumn{2}{|c|}{\textbf{Falcon 40B Instruct Few-Shot prompt}} \\
\hline
\textbf{Template for PPLM Prompts and OWT neutral prompts:} & System: You are a text generator. Reply with only one turn.\par User: Write a fluent English text starting with "The new research reveals relatively recent tectonic activity connected to" with \{positive or science/technology or "lunar", "cooling"\} \{sentiment or topic or keywords\} \par Falcon: The new research reveals relatively recent tectonic activity connected to the long-lived cooling and associated contraction of the lunar interior. \par User: Write a fluent English text starting with "This test article is a quarter the size of" with \{negative or business or "2016", "company", "sail"\} \{sentiment or topic or keywords\} \par Falcon: This test article is a quarter the size of the sail the company plans to fly in 2016.\par User: Write a fluent English text starting with "\{text\}" with a \{control attribute value\} \{control attribute\}\par \\
\hline
\textbf{Template for STS benchmark test and Cloze 2018 test:} & System: You are a story writer. Reply with only one turn. \par User: Write a fluent English story starting with "\{text\}" with a \{control attribute value\} \{sentiment or topic or keywords\} \par Falcon: Jack is tired of writing emails on his tablet computer. He decides he needs to invest in a laptop. He researches different laptops until he has found the perfect one. Finally he buys it. Now Jack actually enjoys replying to emails. \par User: Write a fluent English story starting with "Gertrude was notoriously clumsy and awkward." with a \{negative or sports or "soccer", "ball", "school", "team"\} \{sentiment or topic or keywords\}\par Falcon: Gertrude was notoriously clumsy and awkward. Her friends thought that participating in a sport may help her. She decided to try out for the soccer team at school. She ran out on the field and unexpectedly tripped over the ball. Naturally, she was not accepted to the soccer team. \par User: Write a fluent English story starting with "Jack is tired of writing emails on his tablet computer." with a \{positive or science/technology or "emails", "laptop"\} \{sentiment or topic or keywords\} \\
\hline
\textbf{Multiple-attribute control instruction template:} & User: Write a fluent English \{story or text\} starting with "\{text\}" with \{sentiment value\} sentiment and \{topic value\} topic \\
\hline
\end{tabular}
\caption{\label{tab:temp-few-shot-falc}
Template of the Few-Shot prompt for Falcon 40B Instruct model.}
\end{table*}

\begin{table*}[h!]
\centering\small
\renewcommand{\arraystretch}{1.15}
\begin{tabular}{|>{\raggedright\arraybackslash}p{0.15\textwidth}|>{\raggedright\arraybackslash}p{0.75\textwidth}|}
\hline
\multicolumn{2}{|c|}{\textbf{LLaMa2 70B chat Zero-Shot prompt}} \\
\hline
\textbf{Template for PPLM Prompts and OWT neutral prompts:} & \textless s\textgreater [INST] \textless\textless SYS\textgreater\textgreater \par You are a text generator. Reply with only one turn.\par \textless\textless /SYS\textgreater\textgreater \bigbreak Write a fluent English text starting with "\{text\}" with a \{control attribute value\} \{control attribute\} [/INST] \\
\hline
\textbf{Template for STS benchmark test and Cloze 2018 test:} & \textless s\textgreater [INST] \textless\textless SYS\textgreater \textgreater \par You are a story generator. Reply with only one turn.\par \textless\textless /SYS\textgreater\textgreater \bigbreak Write a fluent English story starting with "\{text\}" with a \{control attribute value\} \{control attribute\} [/INST] \\
\hline
\textbf{Multiple-attribute control instruction template:} & Write a fluent English \{story or text\} starting with "\{text\}" with \{sentiment value\} sentiment and \{topic value\} topic [/INST] \\
\hline
\end{tabular}
\caption{\label{tab:temp-zero-shot-llama}
Template of the Zero-Shot prompt for LLaMa2 70B chat model.}
\end{table*}

\begin{table*}[h!]
\centering\small
\renewcommand{\arraystretch}{1.15}
\begin{tabular}{|>{\raggedright\arraybackslash}p{0.15\textwidth}|>{\raggedright\arraybackslash}p{0.75\textwidth}|}
\hline
\multicolumn{2}{|c|}{\textbf{LLaMa2 70B chat Few-Shot prompt}} \\
\hline
\textbf{Template for PPLM Prompts and OWT neutral prompts:} & \textless s\textgreater [INST] \textless\textless SYS\textgreater\textgreater \par You are a text generator. Reply with only one turn.\par \textless\textless /SYS\textgreater\textgreater \bigbreak Write a fluent English text starting with "The new research reveals relatively recent tectonic activity connected to" with \{positive or science/technology or "lunar", "cooling"\} \{sentiment or topic or keywords\} [/INST] The new research reveals relatively recent tectonic activity connected to the long-lived cooling and associated contraction of the lunar interior. [/INST] Write a fluent English text starting with "This test article is a quarter the size of" with \{negative or business or "2016", "company", "sail"\} \{sentiment or topic or keywords\} [/INST] This test article is a quarter the size of the sail the company plans to fly in 2016. [/INST] Write a fluent English text starting with "\{text\}" with a \{control attribute value\} \{control attribute\} [/INST] \\
\hline
\textbf{Template for STS benchmark test and Cloze 2018 test:} & \textless s\textgreater [INST] \textless\textless SYS\textgreater\textgreater \par You are a story writer. Reply with only one turn.\par \textless\textless /SYS\textgreater\textgreater \bigbreak Write a fluent English story starting with "\{text\}" with a \{control attribute value\} \{sentiment or topic or keywords\} [/INST] Jack is tired of writing emails on his tablet computer. He decides he needs to invest in a laptop. He researches different laptops until he has found the perfect one. Finally he buys it. Now Jack actually enjoys replying to emails. [/INST] Write a fluent English story starting with "Gertrude was notoriously clumsy and awkward." with a \{negative or sports or "soccer", "ball", "school", "team"\} \{sentiment or topic or keywords\} [/INST] Gertrude was notoriously clumsy and awkward. Her friends thought that participating in a sport may help her. She decided to try out for the soccer team at school. She ran out on the field and unexpectedly tripped over the ball. Naturally, she was not accepted to the soccer team. [/INST] Write a fluent English story starting with "Jack is tired of writing emails on his tablet computer." with a \{positive or science/technology or "emails", "laptop"\} \{sentiment or topic or keywords\} \\
\hline
\textbf{Multiple-attribute control instruction template:} & Write a fluent English \{story or text\} starting with "\{text\}" with \{sentiment value\} sentiment and \{topic value\} topic [/INST] \\
\hline
\end{tabular}
\caption{\label{tab:temp-few-shot-llama}
Template of the Few-Shot prompt for LLaMa2 70B chat model.}
\end{table*}

\section{Hyperparameters}
\label{sec:app_hyperparams}
All the scripts and all the library versions used to execute the experiments can be found on our GitHub repo.\footnote{\url{https://github.com/michelalorandi/comparative_study_of_ctg}}

\subsection{CTG Techniques}

We executed all the experiments on our own GPUs.

\paragraph{CTRL} We used \texttt{Salesforce/ctrl} model on HuggingFace both for the tokenier and the model. We set \textit{max length}=256. CTRL was executed on a Nvidia RTXA6000 GPU with 48GB RAM.

\paragraph{C BART} We used the C BART large One-Billion-Words model provided on GitHub\footnote{\url{https://github.com/nlpcode/cbart}}. We set the hyperparameters as follows \textit{decoder chain}=5, \textit{encoder loss type}=0, \textit{max insert label}=1, \textit{temperature}=1, \textit{do sample}=0, \textit{top k}=0, \textit{top p}=0.9, \textit{refinement steps}=10, \textit{max refinement steps}=30, \textit{adaptive}=False, \textit{repetition penalty}=2, \textit{threshold}=0, and \textit{max length}=20. C BART was executed on a Nvidia RTXA6000 GPU with 48GB RAM.

\paragraph{Multi CTG} We used \texttt{bert-base-uncased} and \texttt{gpt2-medium} models on HuggingFace as the pretrained encoder and pretrained decoder, respectively. We also used the checkpoint-30000 provided by the authors\footnote{\url{https://github.com/HappyGu0524/MultiControl}}. The hyperparameters were set as follows: \textit{latent size}=768, \textit{latent num}=1, \textit{sequence length per latent}=20, \textit{variation}=1e-3, \textit{number centers}=1000, \textit{number output centers}=[[1,1,5,1], [10,10,5,1]], \textit{top k}=200, \textit{batch}=5, \textit{max iter}=15, \textit{strategy}=none, \textit{temperature}=50, \textit{SDM reinit}=True, \textit{rp}=1.2, \textit{weight}=\{"default": [2,7,1], "01": [2,4,1], "02": [2,8,1], "03": [3,1,3], "10": [2,12,1], "11": [3,5.5,1], "12": [2,9,1], "13": [3,1,1]\}, and \textit{max length}=50. Multi CTG was executed on a Nvidia RTXA6000 GPU with 48GB RAM.

\paragraph{Prior CTG} We used \texttt{bert-base-uncased} and \texttt{gpt2-medium} models on HuggingFace as the pretrained encoder and pretrained decoder, respectively. We also used the checkpoint-30000 and the configuration files (\textit{config}= generate\_config\_combine.json and \textit{optim config}=generate\_config\_combine\_optim.json) provided by the authors\footnote{\url{https://github.com/HappyGu0524/MultiControl}}. All other hyperparameters were set as follows \textit{latent size}=768, \textit{latent num}=1, \textit{sequence length per latent}=20, \textit{variation}=0, \textit{prior}=True, \textit{flow num}=8, \textit{prior num}=8, \textit{std}=1, \textit{is extend}=True, \textit{is constrained}=True, \textit{weight}=[1,5,1], \textit{optim weight}=[1,1,1], and \textit{max length}=50. Prior CTG was executed on a Nvidia RTXA6000 GPU with 48GB RAM.

\paragraph{CAT-PAW} We used the \texttt{gpt2-medium} model on HuggingFace as the PLM. We set the hyperparameters as follows \textit{stepsize}=0.02, \textit{temperature}=1.0, \textit{top k}=10, \textit{fusion gm scale}=0.9, \textit{fusion kl scale}=0.01, \textit{number iterations}=3, \textit{gradient length}=10000, \textit{number samples}=1, \textit{horizon length}=1, \textit{window length}=0, \textit{decay}=False, \textit{gamma}=1.5, \textit{sample}=True, \textit{activate alter scale}=False, \textit{require origin}=False, \textit{active size}=0.01, \textit{classifier type}=attn, \textit{annotator type}=dis, \textit{loss type}=1, and \textit{max length}=50. CAT-PAW was executed on a Nvidia RTXA6000 GPU with 48GB RAM.

\paragraph{PPLM} We used \texttt{gpt2-medium} model on HuggingFace as the PLM ans we used the models provided by the authors as discriminators\footnote{\url{https://github.com/uber-research/PPLM}}.The common hyperparameters for all control attributes were set as \textit{number samples}=1, \textit{top k}=10, \textit{sample}=True, \textit{gradient length}=10000, \textit{horizon length}=1, \textit{window length}=0, \textit{decay}=False, \textit{kl scale}=0.01, \textit{discriminator weights}=null and \textit{discriminator meta}=null. Regarding sentiment and multiple-attribute control, we set \textit{stepsize}=0.1, \textit{temperature}=0.85, \textit{number iterations}=3, \textit{gamma}=1.5, and \textit{gm scale}=0.9. For topic control we set \textit{stepsize}=0.03, \textit{temperature}=0.9, \textit{number iterations}=10, \textit{gamma}=1.0, \textit{gm scale}=0.95, and \textit{max length}=50. PPLM was executed on a Nvidia RTXA6000 GPU with 48GB RAM.

\paragraph{Falcon 40B Instruct} We used \texttt{tiiuae/falcon-40b-instruct} model on HuggingFace. We set the hyperparameters as \textit{early stopping}=False, \textit{max time}=300, \textit{do sample}=True, \textit{number beans}=1, \textit{number beans groups}=1, \textit{use cache}=True, \textit{temperature}=1.0, \textit{top k}=50, \textit{top p}=0.95, \textit{typical p}=1, \textit{epsilon cutoff}=0, \textit{eta cutoff}=0, \textit{diversity penalty}=0, \textit{repetition penalty}=1, \textit{encoder repetition penalty}=1, \textit{length penalty}=1, \textit{no repeat ngram size}=0, \textit{renormalize logits}=False, \textit{number return sequences}=1, \textit{output attentions}=False, \textit{output hidden states}=False, \textit{output scores}=False, \textit{return dict in generate}=True, and \textit{max length}=100. Falcon was executed on a Nvidia A100 GPU with 80GB RAM.

\paragraph{LLaMa2 70B chat} We used \texttt{meta-llama/Llama-2-70b-chat-hf} model on HuggingFace. We set the hyperparameters as \textit{early stopping}=False, \textit{max time}=300, \textit{do sample}=True, \textit{number beans}=1, \textit{number beans groups}=1, \textit{use cache}=True, \textit{temperature}=1.0, \textit{top k}=50, \textit{top p}=0.95, \textit{typical p}=1, \textit{epsilon cutoff}=0, \textit{eta cutoff}=0, \textit{diversity penalty}=0, \textit{repetition penalty}=1, \textit{encoder repetition penalty}=1, \textit{length penalty}=1, \textit{no repeat ngram size}=0, \textit{renormalize logits}=False, \textit{number return sequences}=1, \textit{output attentions}=False, \textit{output hidden states}=False, \textit{output scores}=False, \textit{return dict in generate}=True, and \textit{max length}=100. LLaMa2 was executed on a Nvidia A100 GPU with 80GB RAM.

\paragraph{GeDi} We used \texttt{gpt2-xl} model on HuggingFace as the frozen PLM and the GeDi sentiment and GeDi topic as GeDi models provided by the authors\footnote{\url{https://github.com/salesforce/GeDi}} for sentiment and topic control, respectively. The hyperparameters were set as follows \textit{pad lens}=null, \textit{temperature}=1.0, \textit{top k}=50, \textit{top p}=1.0, \textit{repetition penalty}=1.2, \textit{repetition penalty scale}=10.0, \textit{pad token id}=0, \textit{do sample}=True, \textit{penalize condition}=True, \textit{discriminator weight}=30.0, \textit{filter p}=0.8, \textit{target p}=0.8, \textit{class bias}=0.0, and \textit{max length}=100. GeDi was executed on a Nvidia RTXA6000 GPU with 48GB RAM.

\paragraph{DisCup} We used \texttt{gpt2-large} model on HuggingFace as the PLM and \texttt{distill\_tuning\_ negative\_(6,6)} and \texttt{distill\_tuning\_positive\_(5,5)} checkpoints provided by the authors\footnote{\url{https://github.com/littlehacker26/Discriminator-Cooperative-Unlikelihood-Prompt-Tuning}} as discriminator for sentiment control. We set the hyperparameters as \textit{template}=(5,5) for positive and (6,6) for negative, \textit{tuning name}=distill\_tuning, \textit{pseudo token}=xxx, \textit{use lm finetune}=False, \textit{lstm dropout}=0.0, \textit{discriminator embedding checkpoint}=null, \textit{prompt pad length}=10, \textit{ranking scope}=50, \textit{temperature}=0.01, \textit{beta}=0.4, and \textit{max length}=20. DisCup was executed on a Nvidia RTXA6000 GPU with 48GB RAM.

\paragraph{DExperts} We used \texttt{gpt2-large} model on HuggingFace as the PLM and the \texttt{finetuned\_ gpt2\_positive} and \texttt{finetuned\_gpt2\_negative} models provided by the authors\footnote{\url{https://github.com/alisawuffles/DExperts}} as experts and anti-experts models. We set the hyperparameters as \textit{model type}=dexperts, \textit{n}=1, \textit{alpha}=0.0, \textit{p}=1.0, \textit{filter p}=0.9, \textit{resume}=False, \textit{use dataset}=True, and \textit{max length}=20. DExperts was executed on a Nvidia RTXA6000 GPU with 48GB RAM.

\paragraph{BOLT} We used \texttt{gpt2-large} model and \texttt{gpt2} tokenizer on HuggingFace and the discriminator checkpoint (\texttt{replaced\_vocab\_roberta\_for\_yelp\_ polarity}) for sentiment control provided by the authors\footnote{\url{https://github.com/launchnlp/BOLT}}. The hyperparameters were set as follows \textit{weight decay}=0.01, \textit{init from vocabulary}=True, \textit{learning rate}=0.025 for sentiment and 0.4 for keywords control, and \textit{max length}=50. BOLT was executed on a Nvidia RTXA6000 GPU with 48GB RAM.

\subsection*{Evaluation Metrics}

\paragraph{SLOR} We used \texttt{gpt2-xl} and \texttt{bigscience/bloom-1b7} models from HuggingFace to compute sentence and unigram probabilities.

\paragraph{Control Effectiveness} To detect sentiment, we used \texttt{distilbert/distilbert-base-uncased-finetuned-sst -2-english}\footnote{\url{https://huggingface.co/distilbert/distilbert-base-uncased-finetuned-sst-2-english}} and \texttt{michelecafagna26/t5-base-finetuned-sst2-sentiment}\footnote{\url{https://huggingface.co/michelecafagna26/t5-base-finetuned-sst2-sentiment}} from HuggingFace, while for topic classification, we used \texttt{textattack/distilbert-base-uncased-ag-news}\footnote{\url{https://huggingface.co/textattack/distilbert-base-uncased-ag-news}} and \texttt{fabriceyhc/bert-base- uncased-ag\_news}\footnote{\url{https://huggingface.co/fabriceyhc/bert-base-uncased-ag_news}}. To compute Word Inclusion Coverage, we used Spacy with \texttt{en\_core\_web\_sm} model to extract lemmas of tokens.

\paragraph{Perplexity} We used the \texttt{evaluate}\footnote{\url{https://huggingface.co/docs/evaluate/en/index}} library of HuggingFace using both \texttt{gpt2-xl} and \texttt{bigscience/bloom-1b7} models.

\section{Scientific artifacts and licensing}
\label{sec:app_license}
CTRL and GeDi are licensed under BSD-3-Claude license. Multi CTG and Prior CTG are licensed under MIT license. PPLM, Falcon 40B Instruct, PPLM Prompts, Yelp review dataset and AGnews dataset are licensed under Apache-2.0 license. LLaMa2 70B chat is licensed under a commercial license.\footnote{\url{https://ai.meta.com/resources/models-and-libraries/llama-downloads/}} BLOOM 1B7 is licensed under RAIL License v1.0.\footnote{\url{https://huggingface.co/spaces/bigscience/license}} C BART, CAT-PAW, DisCup, DExperts, BOLT and Cloze 2018 test were shared without explicitly specified license. Bookcorpus dataset is licensed under Smashwords terms of service.\footnote{\url{https://www.smashwords.com/about/tos}} One-Billion-Words is licensed under CC0: Public Domain license. OWT neutral prompts is licensed under Creative Commons CC0 license (“no rights reserved”). STS benchmark test provides a license for each included dataset.\footnote{\url{http://ixa2.si.ehu.eus/stswiki/index.php/STSbenchmark}} The usage of all listed artifacts is consistent with their licenses.

\section{Reproducibility Checklist for JAIR}

Select the answers that apply to your research -- one per item. 

\subsection*{All articles:}

\begin{enumerate}
    \item All claims investigated in this work are clearly stated. 
    [yes]
    \item Clear explanations are given how the work reported substantiates the claims. 
    [yes]
    \item Limitations or technical assumptions are stated clearly and explicitly. 
    [yes]
    \item Conceptual outlines and/or pseudo-code descriptions of the AI methods introduced in this work are provided, and important implementation details are discussed. 
    [yes]
    \item 
    Motivation is provided for all design choices, including algorithms, implementation choices, parameters, data sets and experimental protocols beyond metrics.
    [yes]
\end{enumerate}

\subsection*{Articles containing theoretical contributions:}
Does this paper make theoretical contributions? 
no

If yes, please complete the list below.

\begin{enumerate}
    \item All assumptions and restrictions are stated clearly and formally. 
    [yes/partially/no]
    \item All novel claims are stated formally (e.g., in theorem statements). 
    [yes/partially/no]
    \item Proofs of all non-trivial claims are provided in sufficient detail to permit verification by readers with a reasonable degree of expertise (e.g., that expected from a PhD candidate in the same area of AI). [yes/partially/no]
    \item
    Complex formalism, such as definitions or proofs, is motivated and explained clearly.
    [yes/partially/no]
    \item 
    The use of mathematical notation and formalism serves the purpose of enhancing clarity and precision; gratuitous use of mathematical formalism (i.e., use that does not enhance clarity or precision) is avoided.
    [yes/partially/no]
    \item 
    Appropriate citations are given for all non-trivial theoretical tools and techniques. 
    [yes/partially/no]
\end{enumerate}

\subsection*{Articles reporting on computational experiments:}
Does this paper include computational experiments? [yes]

If yes, please complete the list below.
\begin{enumerate}
    \item 
    All source code required for conducting experiments is included in an online appendix 
    or will be made publicly available upon publication of the paper.
    The online appendix follows best practices for source code readability and documentation as well as for long-term accessibility.
    [yes]
    \item The source code comes with a license that
    allows free usage for reproducibility purposes.
    [yes]
    \item The source code comes with a license that
    allows free usage for research purposes in general.
    [yes]
    \item 
    Raw, unaggregated data from all experiments is included in an online appendix 
    or will be made publicly available upon publication of the paper.
    The online appendix follows best practices for long-term accessibility.
    [yes]
    \item The unaggregated data comes with a license that
    allows free usage for reproducibility purposes.
    [yes]
    \item The unaggregated data comes with a license that
    allows free usage for research purposes in general.
    [yes]
    \item If an algorithm depends on randomness, then the method used for generating random numbers and for setting seeds is described in a way sufficient to allow replication of results. 
    [yes]
    \item The execution environment for experiments, the computing infrastructure (hardware and software) used for running them, is described, including GPU/CPU makes and models; amount of memory (cache and RAM); make and version of operating system; names and versions of relevant software libraries and frameworks. 
    [yes]
    \item 
    The evaluation metrics used in experiments are clearly explained and their choice is explicitly motivated. 
    [yes]
    \item 
    The number of algorithm runs used to compute each result is reported. 
    [yes]
    \item 
    Reported results have not been ``cherry-picked'' by silently ignoring unsuccessful or unsatisfactory experiments. 
    [yes]
    \item 
    Analysis of results goes beyond single-dimensional summaries of performance (e.g., average, median) to include measures of variation, confidence, or other distributional information. 
    [yes]
    \item 
    All (hyper-) parameter settings for 
    the algorithms/methods used in experiments have been reported, along with the rationale or method for determining them. 
    [yes]
    \item 
    The number and range of (hyper-) parameter settings explored prior to conducting final experiments have been indicated, along with the effort spent on (hyper-) parameter optimisation. 
    [yes]
    \item 
    Appropriately chosen statistical hypothesis tests are used to establish statistical significance
    in the presence of noise effects.
    [no]
\end{enumerate}

\subsection*{Articles using data sets:}
Does this work rely on one or more data sets (possibly obtained from a benchmark generator or similar software artifact)? 
[yes]

If yes, please complete the list below.
\begin{enumerate}
    \item 
    All newly introduced data sets 
    are included in an online appendix 
    or will be made publicly available upon publication of the paper.
    The online appendix follows best practices for long-term accessibility with a license
    that allows free usage for research purposes.
    [NA]
    \item The newly introduced data set comes with a license that
    allows free usage for reproducibility purposes.
    [NA]
    \item The newly introduced data set comes with a license that
    allows free usage for research purposes in general.
    [NA]
    \item All data sets drawn from the literature or other public sources (potentially including authors' own previously published work) are accompanied by appropriate citations.
    [yes]
    \item All data sets drawn from the existing literature (potentially including authors’ own previously published work) are publicly available. [yes]
    \item All new data sets and data sets that are not publicly available are described in detail, including relevant statistics, the data collection process and annotation process if relevant.
    [NA]
    \item 
    All methods used for preprocessing, augmenting, batching or splitting data sets (e.g., in the context of hold-out or cross-validation)
    are described in detail. [yes]
\end{enumerate}

\subsection*{Explanations on any of the answers above (optional):}

NA

\end{document}